\documentclass[journal]{IEEEtran}
\usepackage{cite}
\usepackage{graphicx}
\usepackage{amsmath}
\usepackage{array}
\usepackage{amssymb}
\usepackage{booktabs}
\usepackage{bbding}
\usepackage{subfigure}
\usepackage{textcomp}
\usepackage{todonotes}
\usepackage{gensymb}
\usepackage{xcolor}
\usepackage{color}
\usepackage{url}
\usepackage{multirow}
\usepackage{epsfig}
\usepackage{threeparttable}
\usepackage{algorithmicx,algorithm}
\usepackage{algpseudocode}
\usepackage{bbm}

\def\ie{{\em i.e.}}
\def\eg{{\em e.g.}}
\def\etal{{\em et al.}}

\usepackage{diagbox}
\begin{document}
\title{FM-ViT: Flexible Modal Vision Transformers for Face Anti-Spoofing}
\author{
	Ajian Liu, 
	Zichang Tan, 
	Zitong Yu, 
	Chenxu Zhao, 
	Jun Wan,~\IEEEmembership{Senior Member,~IEEE}, 
	Yanyan Liang \\
	Zhen Lei,~\IEEEmembership{Senior Member,~IEEE}, 
	Du Zhang, 
	Stan Z. Li~\IEEEmembership{Fellow,~IEEE}, 
	Guodong Guo,~\IEEEmembership{Senior Member,~IEEE}
\thanks{Ajian Liu, Jun Wan and Zhen Lei are with the State Key Laboratory of Multimodal Artificial Intelligence Systems (MAIS), Institute of Automation, Chinese Academy of Sciences (CASIA). Beijing, China (e-mail: ajianliu92@gmail.com, \{jun.wan, zhen.lei\}@nlpr.ia.ac.cn).}
\thanks{Zichang Tan and Guodong Guo are with the Institute of Deep Learning, Baidu Research and National Engineering Laboratory for Deep Learning Technology and Application (e-mail: \{tanzichang, guoguodong01\}@baidu.com).}
\thanks{Zitong Yu is with the Great Bay University, Dongguan 523000, China (e-mail: zitong.yu@ieee.org).}
\thanks{Chenxu Zhao is with the Mininglamp Academy of Sciences, Mininglamp Technology, China (e-mail: lonicera.zhao@gmail.com).}
\thanks{Yanyan~Liang and Du~Zhang are with the Macau University of Science and Technology, Macau, China (e-mail: \{yyliang, duzhang\}@must.edu.mo).}
\thanks{Stan Z. Li is with the Westlake University (e-mail: Stan.ZQ.Li@westlake.edu.cn).}
\thanks{\emph{Ajian Liu and Zichang Tan are joint first authors.}}
}
\markboth{}
{Liu \MakeLowercase{~\etal}: FM-ViT: Flexible Modal Vision Transformers for Face Anti-Spoofing}
\maketitle

\begin{abstract}
	The availability of handy multi-modal (\ie, RGB-D) sensors has brought about a surge of face anti-spoofing research. However, the current multi-modal face presentation attack detection (PAD) has two defects: (1) The framework based on multi-modal fusion requires providing modalities consistent with the training input, which seriously limits the deployment scenario. (2) The performance of ConvNet-based model on high fidelity datasets is increasingly limited. In this work, we present a pure transformer-based framework, dubbed the Flexible Modal Vision Transformer (FM-ViT), for face anti-spoofing to flexibly target any single-modal (\ie, RGB) attack scenarios with the help of available multi-modal data. Specifically, FM-ViT retains a specific branch for each modality to capture different modal information and introduces the Cross-Modal Transformer Block (CMTB), which consists of two cascaded attentions named Multi-headed Mutual-Attention (MMA) and Fusion-Attention (MFA) to guide each modal branch to mine potential features from informative patch tokens, and to learn modality-agnostic liveness features by enriching the modal information of own $\texttt{CLS}$ token, respectively. Experiments demonstrate that the single model trained based on FM-ViT can not only flexibly evaluate different modal samples, but also outperforms existing single-modal frameworks by a large margin, and approaches the multi-modal frameworks introduced with smaller FLOPs and model parameters.
\end{abstract}
\IEEEpeerreviewmaketitle

%%%%%%%%% BODY TEXT
\section{Introduction}
\IEEEPARstart{F}{ace} Anti-Spoofing (FAS) aims at protecting face recognition system from various Presentation Attacks (PAs). It has become an increasingly critical concern, including competition organization~\cite{liu2019multi,zhang2020casia,liu20213d,fang2023surveillance} and algorithm design~\cite{Liu2018Learning,yu2020searching,zhang2020face,liu2021face}, due to its wide applications in financial payment, phone unlocking, and access control.
{\tiny \begin{figure}[t]
	\centering
	\includegraphics[width=1.0\linewidth]{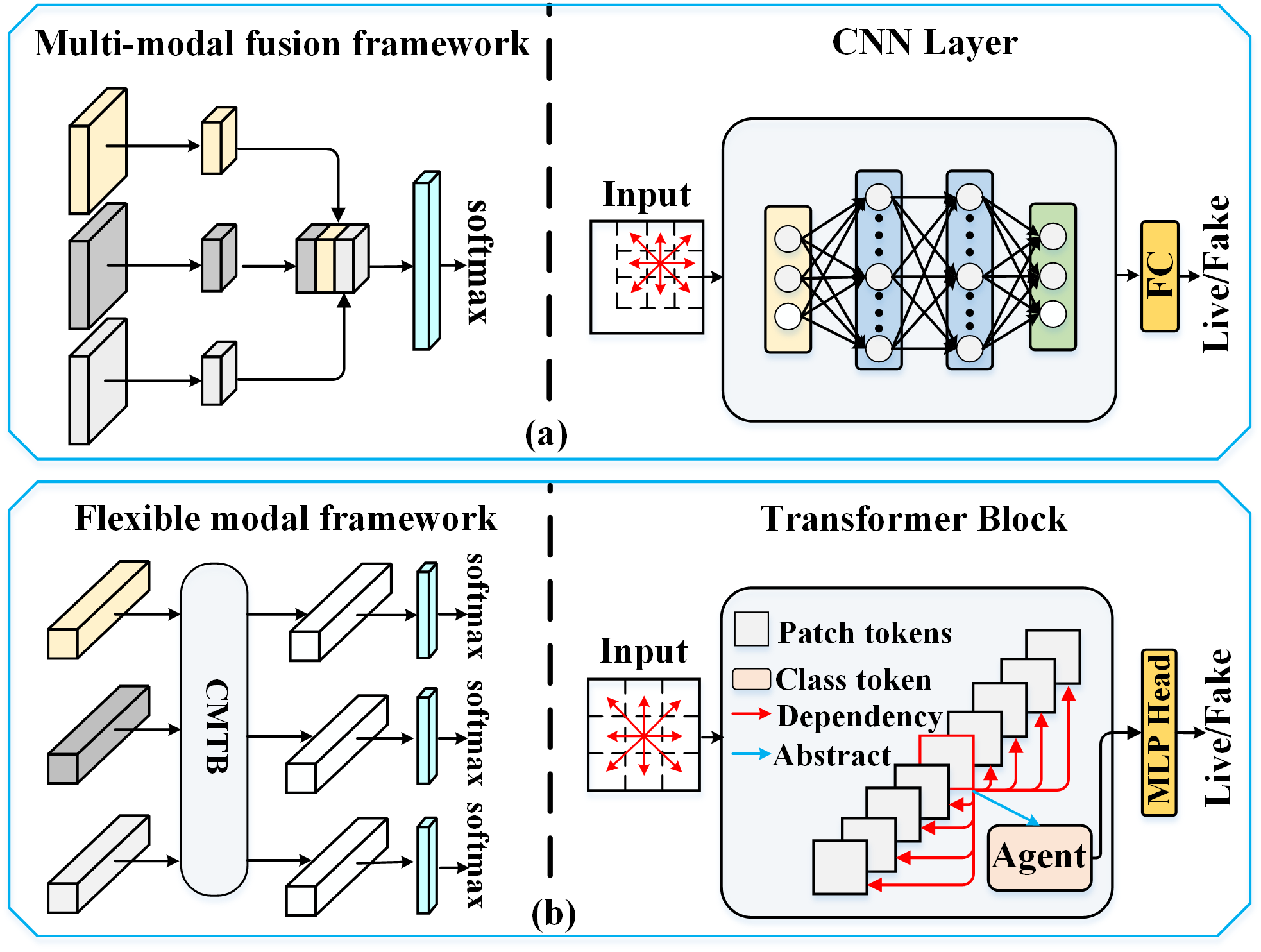}
	\caption{Comparison with existing multi-modal methods. (a) the previous multi-modal fusion methods usually extract features of different modalities based on stacked CNN layers, and then summarize the features with a late (or halfway) fusion strategy for final FAS tasks. (b) Our flexible modal framework is built on multi-branch ViT to realize the function of the flexible modal testing. Further, the CMTB module is introduced to improve the performance of a single-modal system with the help of available multi-modal data.}
	\label{fig:vit_cnn}
\end{figure}
}
With the increasingly advanced presentation attack instruments (PAIs) and acquisition sensors, face presentation attack detection (PAD) algorithms are also expanded from RGB spectrum to multi-spectrum~\cite{zhang2020casia,parkin2019recognizing,george2019biometric}, aim to explore more reliable spoofing traces. Although some recent multi-modal methods can improve the robustness of PAD systems with the help of multi-spectral imaging technology, they require consistent training and test modalities, and lack consideration of the application scenario when a certain modality is missing. See the multi-modal fusion framework in Fig.~\ref{fig:vit_cnn} (a), these methods are usually based on a halfway fusion strategy~\cite{zhang2019dataset}. First, independent branches are used to learn the features of specific modalities, and then these features are summarized at a later stage via the feature concatenation based on a shared branch for the final FAS task. When any modality disappears during testing, these methods would fail to distinguish live vs. fake faces and result in poor performance. Due to hardware cost and space constraints, consistent modal samples cannot always be provided in practical applications, which makes these systems difficult to deploy widely, even if they are robust. Although convolution-based algorithms~\cite{yang2014learn,feng2016integration,li2016original,Patel2016Secure,Liu2018Learning,george2019deep,yu2020searching} remain dominant due to its ability in capturing multiple semantic features, see the CNN layer in Fig.~\ref{fig:vit_cnn} (a), which is enough to distinguish the attacks with obvious spoofing traces~\cite{2010Face,Zhang2012A,Chingovska_BIOSIG-2012}, they are easily confused by high-quality attack samples~\cite{Boulkenafet2017OULU,Liu2018Learning,george2019biometric,CelebA-Spoof,liu2022contrastive} with the increasingly advanced PAIs and acquisition sensors.

In this context, different from a modality fusion framework, our goal is to design a flexible modal framework, as shown in Fig.~\ref{fig:vit_cnn} (b), that can be deployed in any given single-modal testing scenario. Further, in order to improve the performance of a single-modal system with the help of multi-modal data, after learning the individual modal features from different branches, the proposed CMTB guides each branch to learn the potential and modality-agnostic liveness features by summarizing the multi-modal information. Finally, the ideal model is that no matter what modal sample it receives, the output is better than the performance of the model trained with the corresponding modality alone.

Alternatively, Vision Transformer (ViT)~\cite{dosovitskiy2020vit} has demonstrated promising performance on various challenging computer vision tasks. Compared with CNNs in Fig.~\ref{fig:vit_cnn} (b), Transformer encourages non-local computation, captures the global context, and establishes the dependency with a target. We analyze the following three advantages of ViT in solving FAS task: (1) Long-range dependencies. Local spoofing traces, \eg, a paper boundary in print attack, a specular spot in video-replay attack, and specular highlights in 3D mask attack, can establish long-range dependencies with other patch tokens and play its indicating role for a long time. (2) Class token. Usually, these local spoofing traces and other subtle clues related to PAIs are distributed in the whole face in terms of globality and repetition can be summarized by class token, \eg, color distortions~\cite{2015face}, Moir\'e patterns~\cite{liu2020disentangling}, or noise prototypes~\cite{Stehouwer_2020_CVPR}. (3) Multi-headed self-attention. ViT uses a multi-headed attention to capture diverse spoofing traces in parallel. In this case, the multi-headed attention can be viewed as jointly attending to multiple attacks by ensembling multiple attention heads, with each attention head focusing on its specific attention relationship between all patches. (4) Multi-modal compatibility. Compared with CNNs, Transformer can fuse different modalities naturally by embedding them into a common semantic space~\cite{hu2021unit}. After the above analysis, abandoning the CNNs to mine spoofing traces, we design a pure transformer-based framework to resist high fidelity attacks by exploring the spoofing trick from the dependencies between face tokens. To sum up, the main contributions of this paper are summarized as follows:
\begin{itemize}
	\setlength{\itemsep}{1.0pt}
	\item
	To the best of our knowledge, it is the first work to explore Transformer for flexible modal FAS task, where we compare and analyze the feasibility and advantages of Transformer for this community.
	\item
	We present a novel Flexible Modal Vision Transformer (FM-ViT) framework to improve the performance of any single modal FAS system with the help of available multi-modal training data.
	\item
	We develop the Cross-Modal Transformer Block (CMTB) in FM-ViT with two effective attentions, namely MMA and MFA, to achieve the potential region mining and modality-agnostic feature learning respectively.
	\item
	Extensive experiments demonstrate that the proposed FM-ViT can improve the performance of a single-modal system with the help of multi-modal data, with only acceptable FLOPs and model parameters being increased.
\end{itemize}

\section{Related Work}
In this section, we mainly review recent FAS works in three aspects: Single-modal methods, Multi-modal methods, and Vision transformers methods.

{\flushleft \textbf{Single-modal FAS Framework.}}
Since most FAS systems are only equipped with RGB cameras due to the cost and space constraints, the color texture information is an important clue used by previous single-modal methods. Due to early attacks~\cite{2010Face,Zhang2012A,Chingovska_BIOSIG-2012} exposed obvious spoofing traces, the single-modal PAD methods~\cite{yang2014learn,feng2016integration,li2016original,Patel2016Secure} supervised by a simple binary cross-entropy have achieved significant advantages compared with the handcrafted features. However, they treat face PAD as a binary classification, and will highly depend on the biased clues which are not faithful spoof patterns. With the upgrading of PAIs and acquisition sensors, high-quality 2D attacks~\cite{Boulkenafet2017OULU,Liu2018Learning,CelebA-Spoof} have reached visual illegibility. Instead of treating FAS as a simple binary classification, recent single-modal methods~\cite{Liu2018Learning,wang2020deep,yu2020searching,yu2020nasfas,li20203dpc,liu2022disentangling} derive inspiration from physical clues, which are shared by genuine face in any domain, such as depth and material. Liu~\etal~\cite{Liu2018Learning} design a CNN-RNN model to leverage the Depth map and rPPG signal as supervision. Similarly, Wang~\etal~\cite{wang2020deep} takes deep spatial gradient and temporal information to assist depth map regression and Yu~\etal~\cite{yu2020searching} propose a novel frame-level FAS method based on Central Difference Convolution (CDC), which can capture intrinsic detailed patterns via aggregating both intensity and gradient information. Yu~\etal~\cite{yu2020face} treat FAS as a material recognition problem and combine it with classical human material perception, intending to extract discriminative and robust features for FAS task. Instead of using handcrafted binary or pixel-wise labels, Qin~\etal~\cite{qin2021meta} propose a novel Meta-Teacher FAS (MT-FAS) method to train a meta-teacher for supervising PA detectors. To further fully mine the information in local image blocks, \ie, capturing devices and presenting materials, Wang~\etal~\cite{Wang_2022_CVPR} propose PatchNet which reformulates FAS as a fine-grained patch-type recognition problem.

Another works~\cite{Jourabloo2018Face,stehouwer2020noise,liu2020disentangling,zhang2020face} treat the task of FAS as a feature disentangled representation learning. Jourabloo~\etal~\cite{Jourabloo2018Face} solve the face anti-spoofing by inversely decomposing a spoof face into the live face and the spoof noise pattern and then utilizing the spoof noise for classification. Stehouwer~\etal~\cite{stehouwer2020noise} propose a GAN-based architecture to synthesize and identify the noise patterns from seen and unseen medium/sensor combinations. Liu~\etal~\cite{liu2020disentangling} design a Spoof Trace Disentanglement Network (STDN) to disentangle the spoof traces from input faces as a hierarchical combination of patterns at multiple scales. Zhang~\etal~\cite{zhang2020face} propose a novel perspective of face anti-spoofing that disentangles the liveness features and content features from images, and the liveness features are further used for classification under a CNN architecture with multiple appropriate supervisions. There are also some methods~\cite{wang2020cross,Rui2019Multi,wang2019improving,Wangarticle,cai2022learning} that focus on improving the generalization of FAS in unknown domains. In~\cite{Rui2019Multi}, a multi-adversarial deep domain generalization method is proposed to automatically and adaptively learn this generalized feature space shared by multiple source domains. \cite{wang2020cross} proposes an efficient disentangled representation learning for cross-domain face PAD. It consists of a DR-Net and a MD-Net with the purpose of obtaining the live and spoof class distributions and learning domain-independent feature representation from the disentangled features, respectively. Instead of extracting handcrafted features, Cai~\etal~\cite{cai2022learning} propose a learnable network to extract Meta Pattern (MP), and fused them with RGB image through a proposed Hierarchical Fusion Module (HFM). Another works~\cite{wang2020cross,Wangarticle} attempts to improve the generalization capability of PAD into new scenarios via an adversarial domain adaptation or an unsupervised domain adaptation with disentangled representation approach.

However, they tend to overfit attacks seen in the training set, especially when only RGB data is available. In addition, with the popularity of high-fidelity mask attacks~\cite{george2019biometric,9146362,mostaani2020highquality,liu2022contrastive} with more realistic in terms of color, texture, and geometry structure, it is very challenging to mine spoofing traces from the visible spectrum alone.

{\flushleft \textbf{Multi-modal FAS Framework.}}
Multi-modal methods~\cite{zhang2020casia,parkin2019recognizing,george2019biometric,liu2021face,2021Face} have proven to be effective in alleviating the above problems. The motivation for these methods is that indistinguishable fake faces may exhibit quite different properties under the other spectrum.

With the release of several large-scale multi-modal 2D datasets~\cite{steiner2016reliable,zhang2020casia,liu2021casia} and high-fidelity mask datasets~\cite{george2019biometric,heusch2020deep,liu2022contrastive}, exploring multi-modal FAS tasks is of great significance to promote technological progress. For example, Zhang~\etal~\cite{zhang2019dataset} collects a CASIA-SURF dataset with $3$ modalities (\ie, RGB, Depth and NIR) using an Intel RealSense SR300 camera, and proposes a multi-modal multi-scale fusion method for face anti-spoofing. Similarly work, Liu~\etal~\cite{liu2021casia} introduce a CASIA-SURF CeFA dataset, covering $3$ ethnicities, $3$ modalities, $1,607$ subjects, and propose a PSMM-Net~\cite{liu2021casia} to learn the complementary features among different modalities. George~\etal~\cite{george2019biometric} introduce a WMCA database with four channels, \eg, color, depth, near-infrared, and thermal, for face PAD which contains a wide variety of 2D and 3D presentation attacks, and propose a MC-CNN aims to detect sophisticated attacks with multiple channels information. Heusch~\etal~\cite{heusch2020deep} collect a HQ-WMCA database, which can be viewed as an extension of the WMCA~\cite{george2019biometric} database via adding a new sensor acting in the shortwave infrared (SWIR) spectrum. Then, a MC-PixBiS framework is proposed to address the problem of face presentation attack detection using different image modalities. To improve the accuracy of the mask attacks, MLFP~\cite{agarwal2017face}, ERPA~\cite{bhattacharjee2017you}, and 3DMA~\cite{xiao20193dma} also extend the study from visible light to other spectrums, including near-infrared, and thermal spectrums. Yu~\etal~\cite{yu2020multi} combines face depth estimation framework~\cite{Liu2018Learning} and the CDC~\cite{yu2020searching} to fuse multi-modal information and achieved the best results in a FAS competition~\cite{liu2021cross}. Instead of depth and infrared maps, Kong~\etal~\cite{kong2022beyond} devise a novel and cost-effective FAS system based on the acoustic modality, named Echo-FAS, which employs the crafted acoustic signal as the probe to perform face liveness detection.

These fusion methods belong to a halfway fusion strategy, which combines the sub-networks of different modalities at a later stage via the feature map concatenation. However, they require modal input to be consistent with the training phase, which seriously limits the deployment. Recent work~\cite{george2021cross} presents a framework for PAD that uses RGB and depth channels supervised by the proposed cross-modal focal loss (CMFL), which makes it possible to train models using all the available channels and to deploy with a subset of channels. Different from CMFL~\cite{george2021cross} to modulate the loss contribution of each channel, we introduce CMTB module to improve the performance of each branch by fusing multi-modal information.

{\flushleft \textbf{Vision Transformer.}}
Inspired by the success of Transformers in natural language processing (NLP), convolution-free models that only build on transformer blocks have flourished in computer vision. In particular, ViT~\cite{dosovitskiy2020vit} is the first pure transformer architecture replacing all convolutions with self-attention to match or even surpass CNNs in several downstream image tasks. \eg, image classification~\cite{wang2021kvt}, object detection~\cite{carion2020endtoend}, and video classification~\cite{girdhar2019video}. However, it has several drawbacks when compared with CNNs~\cite{wang2021kvt}: large training data, rigid patch division, and single scale. Many variants of vision transformers have also been recently proposed to deal with these problems. DeiT~\cite{touvron2021training} uses distillation for data-efficient training. Swin~\cite{liu2021Swin} produces a hierarchical feature representation by flexibly modeling input at various scales. CrossViT~\cite{chen2021crossvit} constructs a dual-branch vision transformer for learning multi-scale features with a across-attention. In FAS community, ViTranZFAS~\cite{george2021effectiveness} uses the pure ViT to solve the zero-shot anti-spoofing task for the first time. TransRPPG~\cite{yu2021transrppg} proposes a pure rPPG transformer framework for mining the global relationship within MSTmaps for liveness representation. ViTAF~\cite{huang2022adaptive} uses the pure ViT to solve the zero- and few-shot, face anti-spoofing task. Contemporaneous, MA-ViT~\cite{ijcai2022p165} aims to solve flexible modal face anti-spoofing by introducing Modality-Agnostic Transformer Block (MATB), which consists of two stacked attentions named Modal-Disentangle Attention (MDA) and Cross-Modal Attention (CMA), to eliminate modality-related information for each modal sequences and supplement modality-agnostic liveness features from another modal sequences, respectively. Yu~\etal~\cite{yu2023flexiblemodal} establish the first flexible-modal FAS benchmark with the principle `train one for all'. To be specific, with trained multi-modal FAS models, both intra- and cross-dataset testings are conducted on four flexible-modal sub-protocols.

\section{Flexible Modal ViT (FM-ViT)}
In this section, we first introduce the overall architecture of FM-ViT which focuses on how to improve the performance of single-modal FAS system with the help of available multi-modal data. Then, we describe the proposed CMTB module in detail, which consists of Multi-headed Mutual-Attention (MMA) and Fusion-Attention (MFA) to guide each branch to learn potential and modality-agnostic liveness features, respectively.
\begin{figure*}[t]
	\centering
	\includegraphics[width=0.9\linewidth]{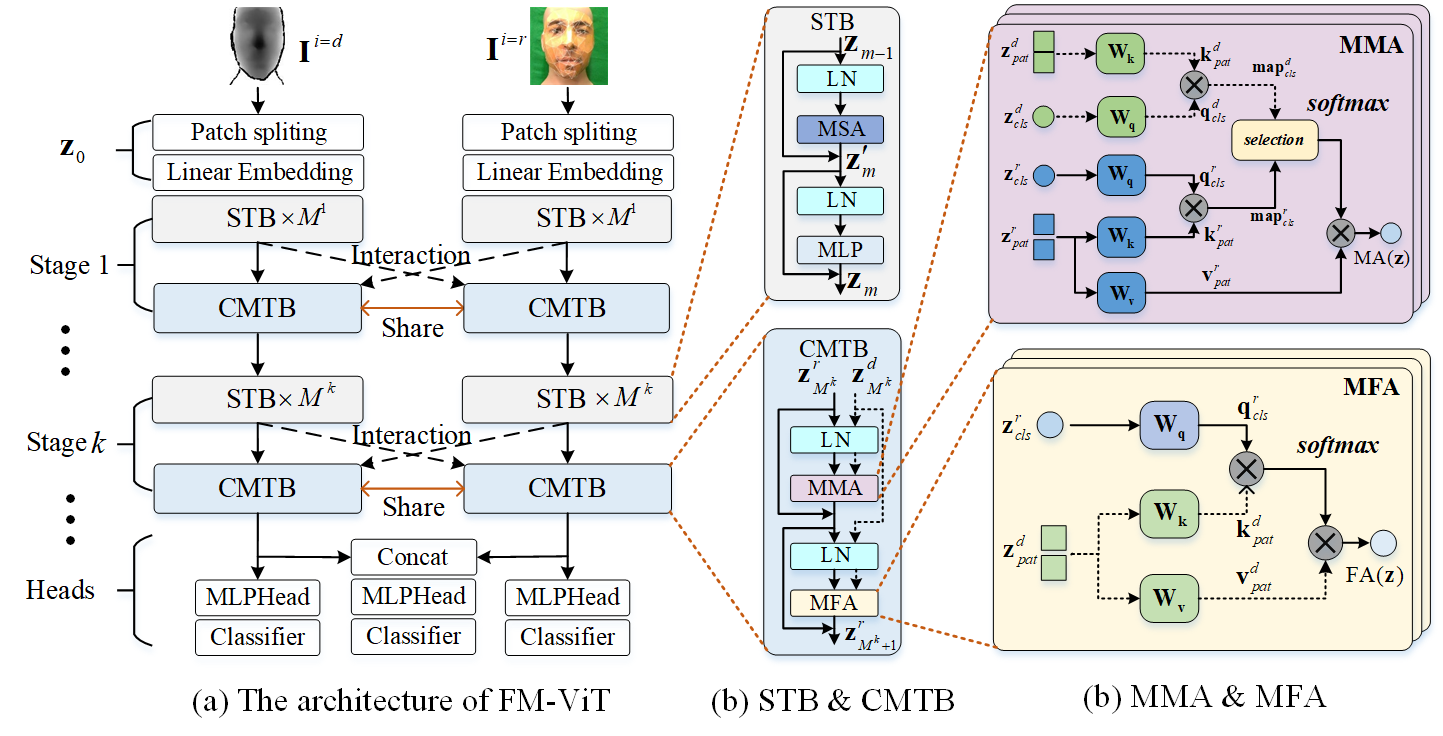}
	\vspace{-0.4cm}
	\caption{(a) The architecture of a Flexible Modal Vision Transformer (FM-ViT). It is built on multiple ViT~\cite{dosovitskiy2020vit} branches and consists of tokenization module (Eq.~\ref{Eq:z0}), transformer encoder and classification heads (Eq.~\ref{Eq:zL}). A completed transformer encoder contains $K$ ``Stage'', in which each ``Stage'' is stacked by $M$ Standard Transformer Blocks (STBs) and a Cross-Modal Transformer Block (CMTB). In each ``Stage'', the CMTB shares the weights (shown by red double arrow line) and receives the output of the previous multi-modal STBs as input (shown by dotted line). (b) STBs~\cite{dosovitskiy2020vit} and CMTB in the $k$-th ``Stage'' of the transformer encoder. The CMTB consists of two cascaded multi-headed mutual-attention (MMA) and multi-headed fusion-attention (MFA). (c) MMA calculates the relevance maps of all modalities to mine the informative patch tokens of its own branch (Eq.~\ref{Eq:mutul-attention}). MFA fuses the modal information of other branches to guide its own branch to learn modality-agnostic liveness features (Eq.~\ref{Eq:fusion-attention}).}
	\label{fig:architecture}
\end{figure*}

\subsection{Overall Framework}
An overview of FM-ViT architecture is depicted in Fig.~\ref{fig:architecture}, which is built on multiple ViT~\cite{dosovitskiy2020vit} branches (each branch corresponds to an input modality) and consists of additional several CMTBs inserted after some specific Standard Transformer Blocks (STBs). Let $\mathbf{I}^{i}$ represents the input at branch $i$, where $i$ can be $r$, $d$, $n$, or $t$ for the RGB, Depth, Near-Infrared (NIR) or Thermal modality. \textbf{Only $\mathbf{I}^{i=r}$ and $\mathbf{I}^{i=d}$ are used to introduce the proposed approach for simplicity.} 

In order to achieve flexible testing of any modality, the FM-ViT assigns a complete ViT branch for each modality, including Tokenization module, Transformer encoder and Classification head. Further, the other modal information is intended to improve the performance of current modality by the proposed CMTB. Specifically, for MMA module, the attention matrix can guide the class token of current modality to pay attention to some ignored patch tokens by the relationship between the class token and patch tokens of other modalities, which can readjust the attention region of current modality by the cues from all other modalities. For MFA module, the class token containing all information of the current modality is used as a query to randomly exchange information with the patch tokens of additional modalities, which can be fused in the class token of the current modality to produce a stronger modality-agnostic representation.

{ \textbf{Multi-Modal Tokenization Module.}}
FM-ViT first splits any modal input $\mathbf{I} \in \mathbb{R}^{H\times W\times C}$ into a sequence of non-overlapping patches $\mathbf{x}_{p} \in \mathbb{R}^{n\times (P^2\cdot C)}$ by a patch splitting module and then linearly projecting patches into tokens $\mathbf{x}_{pat} \in \mathbb{R}^{n\times D}$, where $(H,W,C)$ is the shape of the input image, $(P,P)$ is the resolution of each image patch, $n=HW/P^2$ is the number of resulting patches, and $D$ is vector size through all of FM-ViT layers. Similar to BERT~\cite{devlin2018bert}, a learnable class token ($\texttt{CLS}$) $\mathbf{x}_{cls}=\mathbf{z}_{0,cls} \in \mathbb{R}^{1\times D}$ is concatenated to the sequence of patch tokens, who serves as the image representation (or agent) for classification. And position embeddings $\mathbf{x}_{pos} \in \mathbb{R}^{(n+1)\times D}$ are added to each token embeddings to retain positional information. The tokenization process of sample input is expressed as follows:
\begin{equation}\label{Eq:z0}
	\mathbf{z}_{0}=[\mathbf{x}_{cls}||\mathbf{x}_{pat}]+\mathbf{x}_{pos}, \mathbf{z}_{0}\in \mathbb{R}^{N\times D}, N=n+1.
\end{equation}
where $||$ means token concatenation, and resulting sequence $\mathbf{z}_{0}$ serves as input to the following transformer encoder.

{ \textbf{Transformer Encoder with CMTBs.}}
As shown in Fig.~\ref{fig:architecture}, after tokenization module, sequences $\mathbf{z}^{i}_{0}$ ($i\in \left \{ r,d \right \}$) of all modalities independently pass through $M$ STBs and one CMTB. The complete transformer encoder in FM-ViT contains $K$ ``Stages'' through the above process of stacking. 

See the block detail in Fig.~\ref{fig:architecture}, one STB consists of alternating layers of multi-headed self-attention (MSA) and MLP blocks. Layer normalization (LN) is applied before every block, and residual shortcuts after every block. The CMTB receives the output of the previous multi-modal STBs as input, which consists of two cascaded MMA and MFA. Similar to STBs, LN is applied before every block, and residual shortcuts after every block. However, we do not apply a feed-forward blocks MLP after the MMA and MFA. Specifically, the process of $M^{k}$ STBs in the $k$-th (where $k=1,...,K$) ``Stage'' can be expressed as:
\begin{equation}\label{Eq:stb}
	\begin{split}	
		\mathbf{z'}_{m}&=\mathrm{MSA}(\mathrm{LN}(\mathbf{z}_{m-1}))+\mathbf{z}_{m-1}, m=1,...,M^{k}, \\
		\mathbf{z}_{m}&=\mathrm{MLP}(\mathrm{LN}(\mathbf{z'}_{m}))+\mathbf{z'}_{m}, m=1,...,M^{k}.  \\
	\end{split}
\end{equation}
where the MLP contains two-layer multilayer perceptron. The process of a following CMTB can be expressed as:
\begin{equation}\label{Eq:cmtb}
	\begin{split}		
		\mathbf{z'}^{r}_{M^{k}}&=\mathrm{MMA}(\mathrm{LN}(\mathbf{z}^{r}_{M^{k}}), \mathrm{LN}(\mathbf{z}^{d}_{M^{k}}))+\mathbf{z}^{r}_{M^{k}},\\
		\mathbf{z}^{r}_{M^{k}+1}&=\mathrm{MFA}(\mathrm{LN}(\mathbf{z'}^{r}_{M^{k}}), \mathrm{LN}(\mathbf{z}^{d}_{M^{k}}))+\mathbf{z'}^{r}_{M^{k}}.  \\
	\end{split}
\end{equation}
where $\mathbf{z}^{r}_{M^{k}}$ and $\mathbf{z}^{r}_{M^{k}+1}$ are the outputs of STBs and CMTB for $r$ modality, respectively. Another modal sequence $\mathbf{z}^{d}_{M^{k}}$ participates in the training as inputs of MMA and MFA at the same time. The same procedure is performed for $\mathbf{z}^{d}_{M^{k}+1}$ by simply swapping the index $r$ and $d$.

{ \textbf{Flexible-Modal Classification Heads.}}
To flexibly deploy the proposed framework in the devices with any modal sensor, our model is required to be trained on all the available modalities and tested on any sub-modal sets. In this work, we provide a classification head for each modal branch and a joint classification head for the combined multi-branch sequences to meet the requirements of multi-functional testing, which are supervised by independent binary cross-entropy (BCE). The output sequences at the $K$-th ``Stage'' of transformer encoder $\mathbf{z}^{r}_{K,cls}$ and $\mathbf{z}^{d}_{K,cls}$ are served as the agents of tokens $\mathbf{x}^{r}$ and $\mathbf{x}^{d}$ for classification. The total loss $L^{total}$ to minimize is given as:
\begin{equation}\label{Eq:zL}
	\begin{split}		
		L^{i} &=\mathrm{BCE}(\mathrm{MLP}(\mathrm{LN}(\mathbf{z}_{K,cls}^{i})), y), i\in \left \{ r,d \right \}, \\
		L^{joint} &=\mathrm{BCE}(\mathrm{MLP}(\mathrm{LN}(\mathbf{z}_{K,cls}^{r}||\mathbf{z}_{K,cls}^{d})), y), \\
		L^{total} &=L^{r}+L^{d}+L^{joint}.
	\end{split}
\end{equation}
where $||$ means sequences concatenation along the vector dimension, and the classification head is implemented by MLP with a single linear layer. $y$ is the ground truth ($y=0$ for attack and $y=1$ for bonafide) for sample $\mathbf{I}$.

\subsection{Cross-Modal Transformer Block (CMTB)}
\begin{figure}[t]
	\centering
	\includegraphics[width=1.0\linewidth]{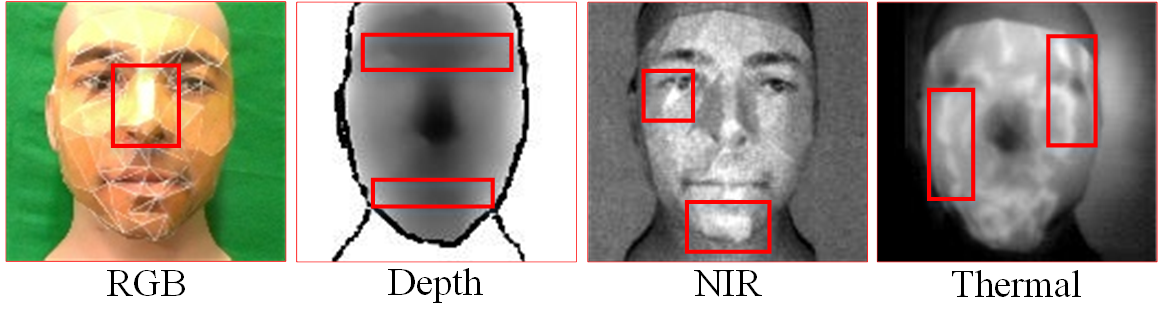}
	\vspace{-0.6cm}
	\caption{A 4-modality PaperMask sample in WMCA. The red box indicates the distribution area of spoofing traces.}
	\label{fig:4_modalities}
\end{figure}

As shown in Fig.~\ref{fig:4_modalities}, for a 4-modality PaperMask sample in WMCA~\cite{george2019biometric}: (1) RGB data has rich appearance details, including the paper creases and the paper reflection spot at the nose. (2) Depth data is sensitive to the distance between the image plane and the corresponding face. However, the paper mask lacks depth change at the eyes and mouth. (3) The eyes and chin of the paper mask expose obvious folding marks in NIR data. (4) The thermal data measures the heat radiated by the face. The paper mask shows uneven heat distribution in the cheek due to occlusion. To sum up, we can conclude that the same attack shows different spoofing cues in different modalities, which are distributed in different face regions.

In this work, the CMTB aims to guide any modal branch to mine its informative patch tokens by the indication of other modalities, and to learn the modality-agnostic features through the developed MMA and MFA respectively.

\iffalse
{\flushleft \textbf{Self-Attention.}}  % $h$ heads
A standard $\mathbf{qkv}$ self-attention (SA) first maps any input sequence $\mathbf{z} \in \mathbb{R}^{N\times D}$ to three matrices of queries $\mathbf{q}$, keys $\mathbf{k}$ and values $\mathbf{v}$ through learnable parameters $\mathbf{W}_{q}$, $\mathbf{W}_{k}$ and $\mathbf{W}_{v}$, respectively. Then, we compute the dot products of the query with all keys, divide each by $\sqrt{D/h}$, and apply a softmax function to obtain the weights. Finally, the final output of self-attention is a weighted sum over all values, expressed as follows:
\begin{equation}\label{Eq:self-attention}
	\begin{split}		
		[\mathbf{q,k,v}]&=[\mathbf{z}\mathbf{W}_{q}, \mathbf{z}\mathbf{W}_{k}, \mathbf{z}\mathbf{W}_{v}], \mathbf{W}\in \mathbb{R}^{D\times (D/h)}, \\
		\textbf{A}&=\mathrm{softmax}(\mathbf{qk}^{T}/\sqrt{D/h}), \textbf{A}\in \mathbb{R}^{N\times N}, \\
		\mathrm{SA}(\textbf{z})&=\textbf{A}\cdot \textbf{v}.
	\end{split}
\end{equation}
where $D$ and $h$ are the embedding dimension and number of heads. The attention matrix \textbf{A} is essentially a relevance map, whose each row corresponds to a link for each token given the other tokens.
\fi

{ \textbf{Multi-headed Mutual-Attention (MMA).}}
How to determine the informative patch tokens is the primary task in MMA. By analyzing the attention matrix~\cite{dosovitskiy2020vit}, which is essentially a relevance map, whose each row corresponds to a link for each token given the other tokens. Therefore, the relevance map that corresponds to the $\texttt{CLS}$ token links each of the tokens to the $\texttt{CLS}$ token, and the strength of this link can be intuitively considered as an indicator of the contribution of each token to the classification~\cite{chefer2020transformerInterpretability}.
\begin{figure}[t]
	\centering
	\includegraphics[width=1.0\linewidth]{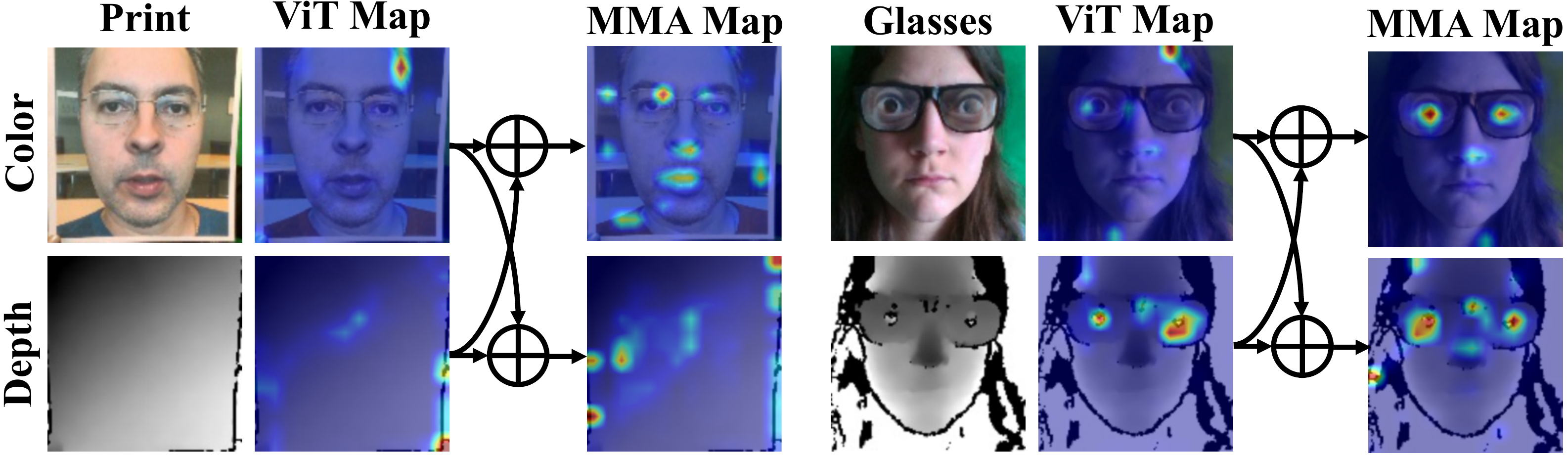}
	\caption{Visualization of classification features based on the vanilla Small-ViT and inserted MMA. Print and Glasses samples are from WMCA~\cite{george2019biometric} dataset. `$\oplus$' means attention matrix accumulation operation.}
	\label{fig:MMA-MAP}
\end{figure}

To corroborate the effect of MMA, we use the vanilla ViT and ViT inserted into the MMA to train the attentions of each layer on the training set of WMCA~\cite{george2019biometric} respectively, and visualize the response map of the final classification feature on the face area on the testing set.  It can be seen from Fig.~\ref{fig:MMA-MAP} that when there is no MMA, the feature response of color modality focuses on the overconfidence area, which leads to the wrong classification of samples, such as the color details in high-definition Print attack and the live face areas that are not covered by funny eyes glasses. After adding MMA, with the guidance of depth modality, it further mining the spoofing details that are easily ignored in the color modality, such as the paper boundary and facial depth in Print attack, and the deliberate funny eyes glasses.

See from the MA module in Fig.~\ref{fig:architecture}, for any modal sequence $\mathbf{z}$, as shown in Eq.~\ref{Eq:mutul-attention}, we first compute $\mathbf{q}_{cls}$ of $\texttt{CLS}$ token, $\mathbf{k}_{pat}$ and $\mathbf{v}_{pat}$ of patch tokens through three learnable parameters $\mathbf{W}_{q}$, $\mathbf{W}_{k}$ and $\mathbf{W}_{v}$, respectively. Then, we compute the dot products of the $\mathbf{q}_{cls}$ with $\mathbf{k}_{pat}$, divide each by $\sqrt{D/h}$, and apply a softmax function to obtain the relevance map $\mathbf{map}_{cls}$. At the same time, we identify the informative patch tokens based on the strong links in $\mathbf{map}_{cls}$ when the modal sequence is classified correctly, which is completed by a threshold function $\Gamma_{\lambda }(\cdot )$. In which $\Gamma_{\lambda}(\cdot )$ aims to find the corresponding patch tokens by thresholding the relevance map to keep $\lambda (\in  [0,1])$ proportional mass. It outputs a mask matrix $\textbf{M}$ with values of 1 and 0, where 1/0 means to retain/discard the patch tokens of the corresponding position. In order to mine informative patch tokens in the current modal sequence, we take the strong links from other modal sequences as indicators to retain the patch tokens discarded by the current modal sequence. As shown in Eq.~\ref{Eq:mutul-attention}, this process can be completed in sequence by accumulating the mask matrix $\textbf{M}^{i}(i\in \left \{ r,d \right \})$ and a selection function $\Gamma'_{\textbf{M}}(\cdot)$ based on the index position of a given matrix $\textbf{M}^{i}$. Finally, we successively obtain the weights by a softmax function to redefined $\mathbf{map}_{cls}$ and the output of MA by the weighted sum over all values of patch tokens $\mathbf{v}_{pat}$. This process is denoted as:
\begin{equation}\label{Eq:mutul-attention}
	\begin{split}		
		[\mathbf{q}_{cls},\mathbf{k}_{pat},\mathbf{v}_{pat}]&=[\mathbf{z}_{cls}\mathbf{W}_{q}, \mathbf{z}_{pat}\mathbf{W}_{k}, \mathbf{z}_{pat}\mathbf{W}_{v}], \\
		\mathbf{map}_{cls}&=\mathbf{q}_{cls}\mathbf{k}_{pat}^{T}/\sqrt{D/h}, \\
		\textbf{M}^{i}&=\Gamma_{\lambda }(\mathbf{map}^{i}_{cls}), \textbf{M}=\textbf{M}^{r}+ \textbf{M}^{d}, \\
		\textbf{A}&=\mathrm{softmax}[\Gamma'_{\textbf{M}}(\mathbf{map}_{cls})], \\
		\mathrm{MA}(\textbf{z})&=\textbf{A}\cdot \mathbf{v}_{pat}.
	\end{split}
\end{equation}
where $D$ and $h$ are the embedding dimension and number of heads, respectively. $\mathbf{W}_{q}$, $\mathbf{W}_{k}$ and $\mathbf{W}_{v}$ $\in \mathbb{R}^{D\times (D/h)}$. $\Gamma'_{\textbf{M}}(\cdot)$ is a selection function defined as:
\begin{equation}\label{Eq:where_operator}
	\begin{split}
		\Gamma'_{\textbf{M}}(\mathbf{A})=\left\{\begin{matrix}
			\mathbf{A}_{a,b}, \textbf{M}_{a,b}>0 & \\ 
			-\infty, \textbf{M}_{a,b}=0 & 
		\end{matrix}\right.	
	\end{split}
\end{equation}
Similar to~\cite{wang2021kvt}, we discard the noise tokens by setting the attention values in $\mathbf{map}_{cls}$ to small enough constant. MMA is an extension of MA in which we run $h$ mutual-attention operations in parallel. As shown in Fig.~\ref{fig:architecture} and Eq.~\ref{Eq:cmtb}, we apply a residual shortcuts after the MMA and concatenate with patch tokens $\mathbf{z}_{pat}$ to obtain a new sequences $\mathbf{z'}$:
\begin{equation}\label{Eq:mma_z'}
	\begin{split}		
		\mathbf{z'}_{cls}=\mathrm{MMA}(\textbf{z})+\mathbf{z}_{cls}, \mathbf{z'} &= [\mathbf{z'}_{cls}||\mathbf{z}_{pat}].
	\end{split}
\end{equation}
where $||$ means token concatenation, the output sequences $\mathbf{z'}$ will be used as the input sequences of the next stage.

Compared to self-attention~\cite{dosovitskiy2020vit} that computes the relevance maps that correspond to all tokens and the attention matrix on all query-key pairs, there are two differences in mutual-attention: (1) we only calculate the relevance map that corresponds to $\texttt{CLS}$ token to mine potential patch tokens. (2) we only select the informative patch tokens from the keys for each query to compute the attention map. Thus, the performance is improved by retaining informative patch tokens, and the training process is accelerated by eliminating noisy tokens (\ie, clutter background and occlusion).

{ \textbf{Multi-headed Fusion-Attention (MFA).}}
How to effectively fuse multi-modal information to produce a stronger modality-agnostic representation is the primary task in MFA. Due to the $\texttt{CLS}$ token being essentially an image agent, which summarizes all the patch tokens for prediction, inspired by cross-attention~\cite{chen2021crossvit}, we develop a simple yet effective multi-modal fusion strategy, which uses the $\texttt{CLS}$ token for each modal sequence as a query to exchange information with patch tokens of other modal sequences. The MFA is inspired by the self-supervision, but has the following differences: (1) Input form. Different from the self-attention, which takes a single modal sample as input, the input the MFA is a pair of samples of different modalities, \ie, $\mathbf{I}^{r}$ and $\mathbf{I}^{d}$, where the query and key/value are from patches of $\mathbf{I}^{r}$ and $\mathbf{I}^{d}$ respectively. (2) Fusion strategy. The purpose of self-attention is to mine the relationship between all the tokens (including $\texttt{CLS}$ token and patch tokens) in the input sample, while the MFA is to fuse the other modal information for current modal sequence by calculating the dependency between the $\texttt{CLS}$ token (from current modality) and the patch tokens (from another modality).

As for the MFA in Fig.~\ref{fig:architecture}, we first compute queries $\mathbf{q}^{r}_{cls}$ with $\texttt{CLS}$ token of modal sequence $\mathbf{z}^{r}$, and keys $\mathbf{k}^{d}_{pat}$, values $\mathbf{v}^{d}_{pat}$ with patch tokens from another modal sequence $\mathbf{z}^{d}$, respectively. Similar to self-attention~\cite{dosovitskiy2020vit} in Eq.~\ref{Eq:fusion-attention}, the attention function $\mathbf{A}$ is computed on the set of queries $\mathbf{q}^{r}_{cls}$ simultaneously with all keys $\mathbf{k}^{d}_{pat}$. Finally, the outputs of FA is a weighted sum over all values $\mathbf{v}^{d}_{pat}$, denoted as:
\begin{equation}\label{Eq:fusion-attention}
	\begin{split}		
		[\mathbf{q}^{r}_{cls},\mathbf{k}^{d}_{pat},\mathbf{v}^{d}_{pat}]&=[\mathbf{z}^{r}_{cls}\mathbf{W}_{q}, \mathbf{z}^{d}_{pat}\mathbf{W}_{k}, \mathbf{z}^{d}_{pat}\mathbf{W}_{v}], \\
		\textbf{A}=\mathrm{softmax}&(\mathbf{q}^{r}_{cls}(\mathbf{k}^{d}_{pat})^{T}/\sqrt{D/h}), \textbf{A}\in \mathbb{R}^{1\times n}, \\
		\mathrm{FA}(\textbf{z})&=\textbf{A}\cdot \textbf{v}^{d}_{pat}.
	\end{split}
\end{equation}
where $\mathrm{FA} (\cdot)$ receives all modal sequences $\mathbf{z}$ as inputs and outputs the $\mathbf{z}^{r}_{cls}$ of the current training sequence $\mathbf{z}^{r}$. Similar to MMA, we also apply a residual shortcuts after the MFA and obtain $\mathbf{z'}^{r}_{cls}$, and concatenate with patch tokens $\mathbf{z}^{r}_{pat}$ to obtain a new sequences $\mathbf{z'}^{r}$ (similar way for sequence $\mathbf{z'}^{d}$). This process is denoted in Eq.~\ref{Eq:mfa_z'}:
\begin{equation}\label{Eq:mfa_z'}
	\begin{split}		
		\mathbf{z'}^{r}_{cls}=\mathrm{MFA}(\textbf{z})+\mathbf{z}^{r}_{cls}, \mathbf{z'}^{r} &= [\mathbf{z'}^{r}_{cls}||\mathbf{z}^{r}_{pat}].
	\end{split}
\end{equation}
where $||$ means token concatenation, the output sequences $\mathbf{z'}^{r}$ will be used as the input sequences of the next stage. Since the $\texttt{CLS}$ token already learns abstract information among all patch tokens in its own modality, interacting with the patch tokens of other modalities can enrich own multi-modal information. In other words, MFA guides the any branch to learn modality-agnostic liveness features by enriching the modal information of own $\texttt{CLS}$ token.

\section{Experiments}
\begin{table*}
	\centering
	\caption{The results on MmFA. A large TPR(\%) and a lower ACER (\%) indicate better performance. Best results are bolded.}
	\resizebox{0.9\linewidth}{!}{
		\begin{tabular}{|c|ccc|c|c|c|}
			\hline
			\multirow{2}{*}{Method} & \multicolumn{3}{c|}{TPR}                                                    & \multirow{2}{*}{APCER} & \multirow{2}{*}{BPCER} & \multirow{2}{*}{ACER} \\ \cline{2-4}
			& \multicolumn{1}{c|}{@FPR=$10^{-2}$} & \multicolumn{1}{c|}{@FPR=$10^{-3}$} & @FPR=$10^{-4}$ &  &  &  \\ \hline  \hline
			SEF~\cite{zhang2019dataset}                     & \multicolumn{1}{c|}{96.70}     & \multicolumn{1}{c|}{81.80}     & 56.80     & 3.80                   & 1.00                   & 2.40                  \\ \hline
			MS-SEF~\cite{zhang2020casia}                  & \multicolumn{1}{c|}{99.70}     & \multicolumn{1}{c|}{97.40}     & 92.40      & 1.90                   & \textbf{0.10}                   & 1.00                  \\ \hline
			% Feather~\cite{2019FeatherNets}                 & \multicolumn{1}{c|}{99.95}     & \multicolumn{1}{c|}{99.83}     & 98.14     & 0.12                   & 0.14                   & 0.13                  \\ \hline
			% ReadSense~\cite{Shen_2019_CVPR_Workshops}               & \multicolumn{1}{c|}{100.00}    & \multicolumn{1}{c|}{99.94}     & 99.80     & 0.19                   & 0.01                   & 0.10                  \\ \hline
			VisionLabs~\cite{2019Recognizing}              & \multicolumn{1}{c|}{\textbf{99.98}}     & \multicolumn{1}{c|}{\textbf{99.95}}     & \textbf{99.87}     & \textbf{0.01}                   & 0.15                   & \textbf{0.08}                  \\ \hline
			ViT                     & \multicolumn{1}{c|}{87.58}     & \multicolumn{1}{c|}{63.09}     & 27.05     & 3.94                   & 4.48                   & 4.21                  \\ \hline
			FM-ViT             & \multicolumn{1}{c|}{99.83}    & \multicolumn{1}{c|}{99.13}    & 98.23     & 0.39     & 0.50     & 0.45    \\ \hline
		\end{tabular}
	}
	\label{tab:casiasurf_result}
\end{table*}
\begin{table}[]
	\centering
	\caption{Evaluation results (\%) on the Protocol 1, 2, and 4 of CeFA dataset.} %Note that a lower ACER value indicates better performance. The best results are bolded.
	\resizebox{1.0\linewidth}{!}{
		\begin{tabular}{|c|c|c|c|c|}
			\hline
			Pro.               & Method  & APCER(\%)     & BPCER(\%)     & ACER(\%)      \\ \hline  \hline
			\multirow{4}{*}{1} & PSMM~\cite{liu2021casia}    & 2.40$\pm$0.60 & 4.60$\pm$2.30 & 3.50$\pm$1.30 \\ \cline{2-5} 
			& ViT     & 1.42$\pm$0.51     & 1.58$\pm$1.88     & 1.50$\pm$0.77     \\ \cline{2-5} 
			%& MA-ViT  & 1.45$\pm$1.75     & 0.75$\pm$0.43     & 1.10$\pm$1.09     \\ \cline{2-5} 
			& FM-ViT  & \textbf{1.29$\pm$1.21}     & \textbf{0.67$\pm$0.95}     & \textbf{0.98$\pm$0.31}     \\ \hline
			\multirow{4}{*}{2} & PSMM~\cite{liu2021casia}    & 7.70$\pm$9.00 & 3.10$\pm$1.60 & 5.40$\pm$5.30 \\ \cline{2-5} 
			& ViT     & 2.82$\pm$1.20     & 1.25$\pm$0.59     & 1.67$\pm$0.83     \\ \cline{2-5} 
			%& MA-ViT  & 0.12$\pm$0.08     & 0.09$\pm$0.12     & 0.10$\pm$0.01     \\ \cline{2-5} 
			& FM-ViT  & \textbf{0.46$\pm$0.09}     & \textbf{1.08$\pm$0.83}     & \textbf{0.30$\pm$0.07}     \\ \hline
			\multirow{7}{*}{4} & PSMM~\cite{liu2021casia}    & 7.80$\pm$2.90 & 5.50$\pm$3.00 & 6.70$\pm$2.20 \\ \cline{2-5} 
			& Hulking~\cite{liu2021cross} & 3.25$\pm$1.98 & 1.16$\pm$1.12 & 2.21$\pm$1.26 \\ \cline{2-5} 
			& Super~\cite{liu2021cross}   & \textbf{0.62$\pm$0.43} & 2.75$\pm$1.50 & 1.68$\pm$0.54 \\ \cline{2-5} 
			& BOBO~\cite{yu2020multi}    & 1.05$\pm$0.62 & 1.00$\pm$0.66 & 1.02$\pm$0.59 \\ \cline{2-5} 
			& ViT     & 3.17$\pm$2.15 & 6.83$\pm$6.08  & 5.00$\pm$2.19 \\ \cline{2-5} 
			%& MA-ViT  & 2.10$\pm$1.47     & 1.17$\pm$0.38     & 1.64$\pm$0.89     \\ \cline{2-5} 
			& FM-ViT  & 0.87$\pm$1.16 & \textbf{0.93$\pm$1.53} & \textbf{0.90$\pm$1.34} \\ \hline
		\end{tabular}
	}
	\label{tab:cefa_result}
\end{table}
\begin{table*}[]
	\centering
	\caption{Comparison of ACER (\%) values on Protocol ``seen" and ``unseen" for the WMCA. Best results are bolded.}
	\scalebox{1.02}{
		\begin{tabular}{|c|c|cccccccc|}
			\hline 
			\multirow{2}{*}{Method} & \multirow{2}{*}{seen} & \multicolumn{8}{c|}{unseen}                                                                                                                                                                                                                      \\ \cline{3-10} 
			&                       & \multicolumn{1}{c|}{Flexiblemask} & \multicolumn{1}{c|}{Replay} & \multicolumn{1}{c|}{Fakehead} & \multicolumn{1}{c|}{Prints} & \multicolumn{1}{c|}{Glasses} & \multicolumn{1}{c|}{Papermask} & \multicolumn{1}{c|}{Rigidmask} & Mean$\pm$Std    \\ \hline \hline
			MC-PixBiS~\cite{george2019deep}                & 1.80                  & \multicolumn{1}{c|}{49.70}        & \multicolumn{1}{c|}{3.70}   & \multicolumn{1}{c|}{0.70}     & \multicolumn{1}{c|}{0.10}   & \multicolumn{1}{c|}{16.00}   & \multicolumn{1}{c|}{\textbf{0.20}}      & \multicolumn{1}{c|}{3.40}      & 10.50$\pm$16.70 \\ \hline
			MCCNN-OCCL-GMM~\cite{george2020learning}          & 3.30                  & \multicolumn{1}{c|}{22.80}        & \multicolumn{1}{c|}{31.40}  & \multicolumn{1}{c|}{1.90}     & \multicolumn{1}{c|}{30.00}  & \multicolumn{1}{c|}{50.00}   & \multicolumn{1}{c|}{4.80}      & \multicolumn{1}{c|}{18.30}     & 22.74$\pm$15.30 \\ \hline
			CMFL~\cite{george2021cross}                    & 1.70                  & \multicolumn{1}{c|}{12.40}        & \multicolumn{1}{c|}{1.00}   & \multicolumn{1}{c|}{2.50}     & \multicolumn{1}{c|}{0.70}   & \multicolumn{1}{c|}{33.50}   & \multicolumn{1}{c|}{1.80}      & \multicolumn{1}{c|}{1.70}      & 7.60$\pm$11.20  \\ \hline
			ViT                     & 2.71                  & \multicolumn{1}{c|}{11.95}        & \multicolumn{1}{c|}{1.44}   & \multicolumn{1}{c|}{3.78}     & \multicolumn{1}{c|}{\textbf{0.00}}   & \multicolumn{1}{c|}{18.02}   & \multicolumn{1}{c|}{0.58}      & \multicolumn{1}{c|}{4.43}      & 5.74$\pm$6.75   \\ \hline
			% MA-ViT                  & 1.45                  & \multicolumn{1}{c|}{9.76}         & \multicolumn{1}{c|}{0.93}   & \multicolumn{1}{c|}{0.55}     & \multicolumn{1}{c|}{0.00}   & \multicolumn{1}{c|}{14.00}   & \multicolumn{1}{c|}{0.00}      & \multicolumn{1}{c|}{1.46}      & 3.81$\pm$5.67    \\ \hline
			FM-ViT                  & \textbf{1.0}                   & \multicolumn{1}{c|}{\textbf{3.56}}         & \multicolumn{1}{c|}{\textbf{0.72}}   & \multicolumn{1}{c|}{\textbf{0.00}}     & \multicolumn{1}{c|}{\textbf{0.00}}   & \multicolumn{1}{c|}{\textbf{12.00}}   & \multicolumn{1}{c|}{0.43}      & \multicolumn{1}{c|}{\textbf{0.73}}      & \textbf{2.49$\pm$4.37}   \\ \hline
		\end{tabular}
	}
	\label{tab:WMCA_results}
\end{table*}
\begin{table}[]
	\centering
	\caption{The HTER (\%) values from the cross-testing between MmFA, CeFA (Protocol 4) and WMCA (Protocol ``seen") datasets.}
	\resizebox{1.0\linewidth}{!}{
		\begin{tabular}{|c|cc|cc|cc|}
			\hline
			\multirow{2}{*}{Method} & \multicolumn{2}{c|}{\begin{tabular}[c]{@{}c@{}}Trained on \\ MmFA\end{tabular}}                                                            & \multicolumn{2}{c|}{\begin{tabular}[c]{@{}c@{}}Trained on \\ CeFA\end{tabular}}                                                            & \multicolumn{2}{c|}{\begin{tabular}[c]{@{}c@{}}Trained on \\  WMCA\end{tabular}}                                                           \\ \cline{2-7} 
			& \multicolumn{1}{c|}{\begin{tabular}[c]{@{}c@{}}Tested on \\ CeFA\end{tabular}} & \begin{tabular}[c]{@{}c@{}}Tested on \\ WMCA\end{tabular} & \multicolumn{1}{c|}{\begin{tabular}[c]{@{}c@{}}Tested on \\ MmFA\end{tabular}} & \begin{tabular}[c]{@{}c@{}}Tested on \\ WMCA\end{tabular} & \multicolumn{1}{c|}{\begin{tabular}[c]{@{}c@{}}Tested on \\ MmFA\end{tabular}} & \begin{tabular}[c]{@{}c@{}}Tested on \\ CeFA\end{tabular} \\ \hline
			ViT                     & \multicolumn{1}{c|}{25.21$\pm$0.75}                                                & 26.50                                                     & \multicolumn{1}{c|}{16.89}                                                     & 33.15                                                     & \multicolumn{1}{c|}{12.61}                                                     & 10.74$\pm$4.13                                                \\ \hline
			MFA-ViT                 & \multicolumn{1}{c|}{24.16$\pm$0.92}                                                & 25.90                                                     & \multicolumn{1}{c|}{15.22}                                                     & 32.11                                                     & \multicolumn{1}{c|}{11.20}                                                     & 10.24$\pm$3.80                                                \\ \hline
			MMA-ViT                 & \multicolumn{1}{c|}{21.51$\pm$0.72}                                                & 22.38                                                     & \multicolumn{1}{c|}{13.41}                                                     & 26.80                                                     & \multicolumn{1}{c|}{9.32}                                                      & 9.34$\pm$3.30                                                 \\ \hline
			FM-ViT                  & \multicolumn{1}{c|}{\textbf{21.43$\pm$1.24}}                                       & \textbf{20.00}                                            & \multicolumn{1}{c|}{\textbf{10.24}}                                            & \textbf{26.34}                                            & \multicolumn{1}{c|}{\textbf{8.45}}                                             & \textbf{8.77$\pm$3.13}                                        \\ \hline
		\end{tabular}
	}
	\label{tab:Cross_testing}
\end{table}
\subsection{Experimental Setup}
{ \textbf{Datasets \& Protocols.}}
We use three commonly used multi-modal and two single-modal FAS datasets for experiments, including CASIA-SURF (MmFA)~\cite{zhang2020casia}, CASIA-SURF CeFA (CeFA)~\cite{liu2021casia}, WMCA~\cite{george2019biometric} OULU-NUPU~\cite{Boulkenafet2017OULU} (OULU) and SiW~\cite{Liu2018Learning}. MmFA~\cite{zhang2020casia} consists of $1,000$ subjects with $21,000$ videos and each sample has $3$ modalities, and provides a intra-testing protocol to evaluate the performance against unknown attack types. CeFA~\cite{liu2021casia} covers $3$ modalities, $1,607$ subjects, and provides five protocols. We select the Protocol 1, 2, and 4 for experiments. WMCA~\cite{george2019biometric} contains a wide variety presentation attacks, which introduces 2 protocols: grandtest protocol emulates the ``seen'' attack scenario and the ``unseen'' protocol evaluates the generalization on an unseen attack. We use Protocol 1 of OULU~\cite{Boulkenafet2017OULU} and SiW~\cite{Liu2018Learning} for cross-testing experiments.

{ \textbf{Test Scenario Settings.}}
We consider two test scenarios. The first is a commonly used setting where the test modalities need to be consistent with the training stage. The second is a flexible modal test scenario, which means the user can provide any single-modal sample. In all scenarios, we evaluate the intra-testing performance based on the provided protocols and the robustness through cross-testing experiments.

{ \textbf{Evaluation Metrics.}}
In intra-testing experiments, Attack Presentation Classification Error Rate (APCER), Bonafide Presentation Classification Error Rate (BPCER), and ACER~\cite{ACER} are used for the metrics. The ACER on the testing set is determined by the Equal Error Rate (EER) threshold on dev sets for MmFA, CeFA, OULU, and the BPCER=$1\%$ threshold for WMCA. TPR(@FPR=$10^{-4}$) is provided for MmFA. For cross-testing experiments, Half Total Error Rate (HTER)~\cite{bengio2004statistical} is adopted as the metric, which computes the average of False Rejection Rate (FRR) and the False Acceptance Rate (FAR), and the threshold computed in dev set using EER criteria.

{ \textbf{Implementation Details.}}
Our models can be freely built on any version of ViT~\cite{dosovitskiy2020vit}. In our experiments, we adopt ViT-S/16 as the backbone through comparative experiments, which means the ``Small'' variant with $K=3$, $M^{1}=2$, $M^{2}=2$ and $M^{3}=4$. Our model is initialized with weights provided by~\cite{dosovitskiy2020vit}, and other newly added layers are randomly initialized. We resize all modal images to $224\times224$ and train all models with $50$ epochs via Adam solver. All models are trained with a batch seize of 8 and an initial learning rate of 0.0001 for all epochs. We set $\lambda=0.5$ in MMA according to conclusion of KVT~\cite{wang2021kvt}. The FM-ViT can be easily extended to more modality, such as RGB (r), Depth (d), NIR (n), Thermal (t). In MMA module, we first calculate the mask matrix \textbf{M} of each modal sequence, which means the location of informative patch tokens in this modality. Then, we force all modalities share an accumulating mask matrix, which can readjust the attention region for each modality by the cues from all other modalities. Therefore, the shared mask matrix is accumulated by four modalities according to $\textbf{M}=\textbf{M}^{r}+ \textbf{M}^{d}+\textbf{M}^{n}+\textbf{M}^{t}$ in Eq.~\ref{Eq:mutul-attention}. In MFA module, for any current sequence (\ie, RGB), we randomly select only one of the remaining modalities (\ie, Depth, NIR, or Thermal) in each iteration for fusion to reduce the computational effort. Therefore, no matter how many modalities are included in one dataset, the MFA is applicable and the calculation amount is fixed. In fact, we found through experiments that there is little difference between the results of fully fusing all modal information in each iteration and the results of randomly selecting a modal fusion.

In the testing stage, for any given single-modal test sample, we input it into all branches to activate CMTB, and report the results by corresponding classification head. For multi-modal test samples, we report the results by joint classification head. In other words, although our system can flexibly test samples with different modalities, we need to know the modal type in advance. To obtain stable experimental results, a statistical test is adopted in all experiments, that is, we report the average of the last five training models as the experimental results.

\subsection{Fixed Modal Scenario Evaluations}  
The fixed modal scenario setting evaluates the fusion ability of our approach to multi-modal information. On the three multi-modal datasets, we compare with baseline method ViT~\cite{dosovitskiy2020vit} that removed the CMTB from FM-ViT, and the previous state-of-the-art (SOTA) methods. 

{ \textbf{Intra-Testing Results.}}  
Because MmFA~\cite{zhang2020casia} is the dataset for holding the multi-modal FAS challenge~\cite{2019Multi}, we compare with the benchmark method SEF~\cite{zhang2019dataset}, multi-scale benchmark method MS-SEF~\cite{zhang2020casia}, and the top one method of the challenge, \ie, VisionLabs~\cite{2019Recognizing}. From the Tab.~\ref{tab:casiasurf_result}, we can observe that the performance of baseline method ViT is worse than SEF due to the lack of modal fusion ability, \ie, $56.80\%$ (SEF) vs. $27.05\%$ (ViT) for TPR@FPR=$10^{-4}$ and $2.40\%$ (SEF) vs. $4.21\%$ (ViT) for ACER. However, when equipped with CMTB, our FM-ViT improves the TPR@FPR=$10^{-4}$ from $27.05\%$ to $98.23\%$, reduces ACER from $4.21\%$ to $0.45\%$, and outperforms MS-SEF by a large margin, \ie, $92.40\%$ (MS-SEF) vs. $98.23\%$ (FM-ViT) for TPR@FPR=$10^{-4}$ and $1.00\%$ (MS-SEF) vs. $0.45\%$ (FM-ViT) for ACER. Finally, the performance of our FM-ViT is worse than that of VisionLabs~\cite{2019Recognizing}. Our analysis has the following two reasons: (1) VisionLabs~\cite{2019Recognizing} adopts additional datasets, \ie, face and gender recognition datasets, and multi-model fusion strategy, \ie, three ResNet-18; (2) Our FM-ViT focuses on the performance of flexible modal testing, and ignores the performance of multi-modal fusion to a certain extent.

Similar to MmFA, CeFA~\cite{liu2021casia} is the dataset for holding another FAS challenge~\cite{liu2021cross} with the Protocol 4. We compare our approach with the benchmark PSMM~\cite{liu2021casia} and the top three methods introduced in the challenge, \ie, BOBO~\cite{yu2020multi}, Super and Hulking~\cite{liu2021cross}. It can be seen from Tab.~\ref{tab:cefa_result} that the performance of baseline ViT is superior to benchmark PSMM due to the pre-training model, while is far worse than the well-designed competition methods due to the lack of modal fusion ability. Such, on Protocol 4, the results of ACER for competition methods are $2.21\%$, $1.68\%$ and $1.02\%$ respectively, which are better than that of ViT ($5.00\%$). Similar conclusions to MmFA, our FM-ViT achieves the best results on all protocols and metrics, \ie, the ACER on Protocol 1, 2 and 4 are $0.98\%$, $0.30\%$ and $0.90\%$, respectively.
%Specifically, the ACER of ViT is $2.00\%$, $3.73\%$ and $1.70\%$ points lower than PSMM on Protocol 1, 2 and 4, and is $2.79\%$, $3.32\%$ and $3.98\%$ points higher than that of Hulking, Super and BOBO on Protocol 4.

To perform a fair comparison with prior methods, only RGB and Depth data in WMCA~\cite{george2019biometric} are used for intra-testing experiments. Tab.~\ref{tab:WMCA_results} presents the comparisons of ACER to the SOTA ConvNet-based methods, including MC-PixBiS~\cite{george2019deep}, MCCNN-OCCL-GMM~\cite{george2020learning}, and CMFL~\cite{george2021cross}. Compared with the previous best results, our FM-ViT achieves significantly better performance with a large margin in ``seen'' and ``unseen'' protocols: $-0.70\%$ for FM-ViT ($1.0\%$) over CMFL ($1.70\%$), and $-5.11\%$ for FM-ViT ($2.49\%$) over CMFL ($7.60\%$), respectively. It is worth noting that FM-ViT noticeably surpasses these methods on two challenging sub-protocols of ``unseen'' protocol: $-8.84\%$ for FM-ViT ($3.56\%$) over CMFL ($12.40\%$) when ``Flexiblemask'' is not seen in the training stage and $-4.0\%$ for FM-ViT ($12.0\%$) over MC-PixBiS ($16.0\%$) when ``Glasses'' is not seen in the training stage. We analyze that the commonality of the two attacks is 3D facial structure, realistic color-texture, and only local regions contain spoofing traces, which are easy to be ignored by the ConvNet-based methods.

{ \textbf{Cross-Testing Results.}}
To evaluate the robustness, we conduct cross-testing experiments between models trained on MmFA, CeFA with Protocol 4, and WMCA with Protocol ``seen''. To demonstrate the contribution of each improvement, we ablate design elements in the proposed FM-ViT, considering baseline ViT and two variants MMA-ViT and MFA-ViT by adding MMA and MFA to the baseline respectively. % We also introduce the previous SOTA ConvNet-based method BOBO~\cite{yu2020multi} as a baseline, which combines the commonly used face depth estimation framework~\cite{Liu2018Learning} and the advanced Central Difference Convolution (CDC)~\cite{yu2020searching} technology.

Tab.~\ref{tab:Cross_testing} lists the results of all methods trained on one dataset and tested on the other two datasets. From these results, it can be seen when trained on MmFA, and tested on CeFA and WMCA datasets, FM-ViT are $3.78\%$ ($21.43\%$ vs. $25.21\%$) and $6.50\%$ ($20.00\%$ vs. $26.50\%$) HTER lower than ViT respectively. Similar conclusions are drawn by comparing training models based on other training data.

Further, it can be concluded that MMA plays a more important role in improving the robustness, compared with MFA. See Tab.~\ref{tab:Cross_testing} for details, when replacing the element of ViT from MFA to MMA, the results are further reduced from left to right successively on 6 test sets. Due to the mismatch of sensors, resolutions, attack types, and settings between different datasets, it is more effective to mine the potential features with the help of MMA against these irrelevant interferences. 
% Similar comparisons for CM-ViT and BOBO, there are $-8.91\%$ ($1.42\%$ vs. $10.33\%$ on MmFA) and $-23.90\%$ ($15.56\%$ vs. $39.46\%$ on WMCA) HTER reductions when the training data is CeFA, and $-5.41\%$ ($1.04\%$ vs. $6.45\%$ on MmFA) and $-6.75\%$ ($2.00\%$ vs. $8.75\%$ on WMCA) HTER reductions when the training data is WMCA. From $11.13\%$, $18.92\%$, $8.33\%$, $28.02\%$, $2.73\%$, and $9.28\%$ to $8.88\%$, $12.34\%$, $5.61\%$, $22.0\%$, $1.50\%$ and $6.10\%$. It demonstrates that FM-ViT is effective in integrating multi-modal information to improve the robustness to unknown data domain. 

\subsection{Flexible Modal Scenario Evaluations}
{ \textbf{Intra-Testing Results.}}
To explore the ability of our approach for modal invariant liveness features learning, we conducted experiments on dataset MmFA~\cite{zhang2020casia}, Protocol 4 of CeFA~\cite{liu2021casia}, and Protocol `seen' of WMCA~\cite{george2019biometric}. Before reporting the results in flexible modal scenarios, we first list the results of the SOTA methods on RGB (R), Depth (D), and IR (I) modalities on each dataset, \ie, MS-SEF~\cite{zhang2020casia} for MmFA dataset, BOBO~\cite{yu2020multi} for CeFA dataset and MC-CNN~\cite{george2020learning} for WMCA dataset. % To avoid the interference of simultaneous training of different modalities, we report the results of independent training of the baseline ViT on each modality.

By comparing the results in Tab.~\ref{tab:flexible_results}, we can draw the following two conclusions: (1) Our FM-ViT achieves better performance than the baseline ViT in each modality, which demonstrates that FM-ViT improves the performance of any single-modal sample with the help of multi-modal information. For example, on the MmFA dataset, the ACER performance of method ViT in RGB, Depth and IR modalities are $16.90\%$, $4.01\%$ and $8.44\%$ respectively, while our FM-ViT reduces the ACER to $12.38\%$, $3.49\%$ and $2.59\%$. Similar conclusions for CeFA and WMCA datasets. Except for RGB modality on CeFA dataset, our algorithm outperform the best performance currently reported on each modality, such as for RGB ($21.00\%$ vs. $12.38\%$), Depth ($3.60\%$ vs. $3.49\%$) and IR ($19.40\%$ vs. $2.59\%$) modalities on MmFA; Depth ($2.73\%$ vs. $2.25\%$) and IR ($10.10\%$ vs. $2.88\%$) modalities on CeFA; RGB ($32.83\%$ vs. $2.87\%$) and Depth ($6.04\%$ vs. $2.32\%$) and IR ($2.51\%$ vs. $2.13\%$) modalities on WMCA. (2) Our FM-ViT reduces the performance gap between different modalities to a certain extent. For example, on WMCA dataset, the performance of method MC-CNN~\cite{george2020learning} on modality RGB, Depth and IR are $32.83\%$, $6.04\%$ and $2.51\%$ respectively, while the results of our FM-ViT are $2.87\%$, $2.32\%$ and $2.13\%$ respectively. Similar conclusions for MmFA and CeFA datasets. We analyze that the MFA module alleviates the bias toward one modality by supplementing the information from other modalities for this current modal sequence.

To verify the advantages of ViT over ResNet in face anti-spoofing task, we select ResNet50~\cite{he2016deep} for comparison, which has similar parameters and flops to the baseline ViT. By comparing the results in Tab.~\ref{tab:flexible_results}, we can observe that for any dataset, the performance of ResNet50 in key indicators is inferior to that of ViT. For example, on MmFA dataset, the ACER of ResNet50 on modality RGB, Depth and IR are $22.95\%$, $4.90\%$ and $21.18\%$ respectively, while the results of ViT are $16.90\%$, $4.01\%$ and $8.44\%$ respectively. The results show that the ViT has more advantages in processing face anti-spoofing task under the condition of using the pre-training model.

\begin{table*}[]
	\centering
	\caption{Comparison of flexible modal results (\%) based on multi-modal datasets. The `SOTA' means the method with public results on the corresponding dataset. R\&D\&I indicates the method receives RGB (R), Depth (D) and IR (I) paired samples as input.}
	%\vspace{-0.3cm}
	\scalebox{0.92}{
		\begin{tabular}{|ccccccccccccccc|}
			\hline
			\multicolumn{1}{|c|}{\multirow{2}{*}{Method}}                                           & \multicolumn{1}{c|}{\multirow{2}{*}{Train}} & \multicolumn{1}{c|}{\multirow{2}{*}{Test}} & \multicolumn{4}{c|}{MmFA}                                                                                         & \multicolumn{1}{c|}{\multirow{2}{*}{}} & \multicolumn{3}{c|}{CeFA (Protocol 4)}                                                               & \multicolumn{1}{c|}{\multirow{2}{*}{}} & \multicolumn{3}{c|}{WMCA (Protocol “seen”)}                     \\ \cline{4-7} \cline{9-11} \cline{13-15} 
			\multicolumn{1}{|c|}{}                                                                  & \multicolumn{1}{c|}{}                       & \multicolumn{1}{c|}{}                      & \multicolumn{1}{c|}{TPR}   & \multicolumn{1}{c|}{APCER} & \multicolumn{1}{c|}{BPCER} & \multicolumn{1}{c|}{ACER}  & \multicolumn{1}{c|}{}                  & \multicolumn{1}{c|}{APCER}       & \multicolumn{1}{c|}{BPCER}      & \multicolumn{1}{c|}{ACER}       & \multicolumn{1}{c|}{}                  & \multicolumn{1}{c|}{APCER} & \multicolumn{1}{c|}{BPCER} & ACER  \\ \hline
			\multicolumn{15}{|c|}{Fixed modal testing}                                                                                                                                                                                                                                                                                                                                                                                                                                                                                                                        \\ \hline
			\multicolumn{1}{|c|}{\multirow{3}{*}{\begin{tabular}[c]{@{}c@{}}SOTA\\ \cite{zhang2020casia,yu2020multi,george2020learning}\end{tabular}}} & \multicolumn{1}{c|}{R}                      & \multicolumn{1}{c|}{R}                     & \multicolumn{1}{c|}{14.60} & \multicolumn{1}{c|}{40.30} & \multicolumn{1}{c|}{1.60}  & \multicolumn{1}{c|}{21.00} & \multicolumn{1}{c|}{\multirow{6}{*}{}} & \multicolumn{1}{c|}{9.96$\pm$5.41}   & \multicolumn{1}{c|}{2.08$\pm$0.88}  & \multicolumn{1}{c|}{6.02$\pm$2.33}  & \multicolumn{1}{c|}{\multirow{6}{*}{}} & \multicolumn{1}{c|}{65.65} & \multicolumn{1}{c|}{0.00}  & 32.83 \\ \cline{2-7} \cline{9-11} \cline{13-15} 
			\multicolumn{1}{|c|}{}                                                                  & \multicolumn{1}{c|}{D}                      & \multicolumn{1}{c|}{D}                     & \multicolumn{1}{c|}{67.30} & \multicolumn{1}{c|}{6.00}  & \multicolumn{1}{c|}{1.20}  & \multicolumn{1}{c|}{3.60}  & \multicolumn{1}{c|}{}                  & \multicolumn{1}{c|}{4.29$\pm$1.37}   & \multicolumn{1}{c|}{1.17$\pm$0.63}  & \multicolumn{1}{c|}{2.73$\pm$0.97}  & \multicolumn{1}{c|}{}                  & \multicolumn{1}{c|}{11.77} & \multicolumn{1}{c|}{0.31}  & 6.04  \\ \cline{2-7} \cline{9-11} \cline{13-15} 
			\multicolumn{1}{|c|}{}                                                                  & \multicolumn{1}{c|}{I}                      & \multicolumn{1}{c|}{I}                     & \multicolumn{1}{c|}{15.90} & \multicolumn{1}{c|}{38.60} & \multicolumn{1}{c|}{0.40}  & \multicolumn{1}{c|}{19.40} & \multicolumn{1}{c|}{}                  & \multicolumn{1}{c|}{19.61$\pm$15.66} & \multicolumn{1}{c|}{0.58$\pm$0.38}  & \multicolumn{1}{c|}{10.10$\pm$7.66} & \multicolumn{1}{c|}{}                  & \multicolumn{1}{c|}{5.03}  & \multicolumn{1}{c|}{0.00}  & 2.51  \\ \cline{1-7} \cline{9-11} \cline{13-15} 
			\multicolumn{1}{|c|}{\multirow{3}{*}{ResNet50}}                                              & \multicolumn{1}{c|}{R}                      & \multicolumn{1}{c|}{R}    & \multicolumn{1}{c|}{1.02}  & \multicolumn{1}{c|}{23.39}  & \multicolumn{1}{c|}{22.50}  & \multicolumn{1}{c|}{22.95} & \multicolumn{1}{c|}{}         & \multicolumn{1}{c|}{25.08$\pm$1.46}  & \multicolumn{1}{c|}{25.06$\pm$1.41} & \multicolumn{1}{c|}{25.07$\pm$1.44} & \multicolumn{1}{c|}{}   & \multicolumn{1}{c|}{6.33}  & \multicolumn{1}{c|}{13.91}  & 10.12
			 \\ \cline{2-7} \cline{9-11} \cline{13-15} 
			\multicolumn{1}{|c|}{}  & \multicolumn{1}{c|}{D}  & \multicolumn{1}{c|}{D}     & \multicolumn{1}{c|}{28.93} & \multicolumn{1}{c|}{2.46}  & \multicolumn{1}{c|}{7.33}  & \multicolumn{1}{c|}{4.90}  & \multicolumn{1}{c|}{}    & \multicolumn{1}{c|}{9.67$\pm$4.54}  & \multicolumn{1}{c|}{8.71$\pm$4.33}  & \multicolumn{1}{c|}{9.19$\pm$4.43}  & \multicolumn{1}{c|}{}  & \multicolumn{1}{c|}{9.50} & \multicolumn{1}{c|}{4.35}  & 6.93  
			\\ \cline{2-7} \cline{9-11} \cline{13-15}  \multicolumn{1}{|c|}{}   & \multicolumn{1}{c|}{I}  & \multicolumn{1}{c|}{I} & \multicolumn{1}{c|}{1.23} & \multicolumn{1}{c|}{26.02}  & \multicolumn{1}{c|}{16.33}  & \multicolumn{1}{c|}{21.18}  	& \multicolumn{1}{c|}{}  & \multicolumn{1}{c|}{6.12$\pm$6.95}   & \multicolumn{1}{c|}{5.13$\pm$2.91}  & \multicolumn{1}{c|}{5.65$\pm$3.25}  & \multicolumn{1}{c|}{}    & \multicolumn{1}{c|}{4.75}  & \multicolumn{1}{c|}{6.96}  & 5.85  \\ \hline
			\multicolumn{1}{|c|}{\multirow{3}{*}{ViT}}                                              & \multicolumn{1}{c|}{R}                      & \multicolumn{1}{c|}{R}                     & \multicolumn{1}{c|}{1.50}  & \multicolumn{1}{c|}{16.64} & \multicolumn{1}{c|}{17.17} & \multicolumn{1}{c|}{16.90} & \multicolumn{1}{c|}{}                  & \multicolumn{1}{c|}{34.74$\pm$5.44}  & \multicolumn{1}{c|}{13.67$\pm$3.25} & \multicolumn{1}{c|}{24.20$\pm$2.34} & \multicolumn{1}{c|}{}                  & \multicolumn{1}{c|}{4.30}  & \multicolumn{1}{c|}{4.35}  & 4.32  \\ \cline{2-7} \cline{9-11} \cline{13-15} 
			\multicolumn{1}{|c|}{}                                                                  & \multicolumn{1}{c|}{D}                      & \multicolumn{1}{c|}{D}                     & \multicolumn{1}{c|}{32.00} & \multicolumn{1}{c|}{4.30}  & \multicolumn{1}{c|}{3.72}  & \multicolumn{1}{c|}{4.01}  & \multicolumn{1}{c|}{}                  & \multicolumn{1}{c|}{8.41$\pm$5.36}   & \multicolumn{1}{c|}{3.83$\pm$3.76}  & \multicolumn{1}{c|}{6.12$\pm$2.91}  & \multicolumn{1}{c|}{}                  & \multicolumn{1}{c|}{5.66}  & \multicolumn{1}{c|}{0.00}  & 2.83  \\ \cline{2-7} \cline{9-11} \cline{13-15} 
			\multicolumn{1}{|c|}{}                                                                  & \multicolumn{1}{c|}{I}                      & \multicolumn{1}{c|}{I}                     & \multicolumn{1}{c|}{20.33} & \multicolumn{1}{c|}{7.15}  & \multicolumn{1}{c|}{9.33}  & \multicolumn{1}{c|}{8.44}  & \multicolumn{1}{c|}{}                  & \multicolumn{1}{c|}{7.90$\pm$6.53}   & \multicolumn{1}{c|}{2.50$\pm$2.65}  & \multicolumn{1}{c|}{5.20$\pm$3.74}  & \multicolumn{1}{c|}{}                  & \multicolumn{1}{c|}{2.94}  & \multicolumn{1}{c|}{1.74}  & 2.34  \\ \hline
			\multicolumn{15}{|c|}{Flexible modal testing}                                                                                                                                                                                                                                                                                                                                                                                                                                                                                                                     \\ \hline
			\multicolumn{1}{|c|}{\multirow{3}{*}{FM-ViT}}                                           & \multicolumn{1}{c|}{R\&D\&I}                      & \multicolumn{1}{c|}{R}                     & \multicolumn{1}{c|}{24.39} & \multicolumn{1}{c|}{8.77}  & \multicolumn{1}{c|}{16.00} & \multicolumn{1}{c|}{12.38} & \multicolumn{1}{c|}{\multirow{3}{*}{}} & \multicolumn{1}{c|}{36.61$\pm$9.51}  & \multicolumn{1}{c|}{5.50$\pm$1.32}  & \multicolumn{1}{c|}{21.06$\pm$4.90} & \multicolumn{1}{c|}{\multirow{3}{*}{}} & \multicolumn{1}{c|}{2.26}  & \multicolumn{1}{c|}{3.48}  & 2.87  \\ \cline{2-7} \cline{9-11} \cline{13-15} 
			\multicolumn{1}{|c|}{}                                                                  & \multicolumn{1}{c|}{R\&D\&I}                      & \multicolumn{1}{c|}{D}                     & \multicolumn{1}{c|}{47.17} & \multicolumn{1}{c|}{5.14}  & \multicolumn{1}{c|}{1.83}  & \multicolumn{1}{c|}{3.49}  & \multicolumn{1}{c|}{}                  & \multicolumn{1}{c|}{2.79$\pm$0.44}   & \multicolumn{1}{c|}{1.71$\pm$1.13}  & \multicolumn{1}{c|}{2.25$\pm$0.36}  & \multicolumn{1}{c|}{}                  & \multicolumn{1}{c|}{2.04}  & \multicolumn{1}{c|}{2.61}  & 2.32  \\ \cline{2-7} \cline{9-11} \cline{13-15} 
			\multicolumn{1}{|c|}{}                                                                  & \multicolumn{1}{c|}{R\&D\&I}                      & \multicolumn{1}{c|}{I}                     & \multicolumn{1}{c|}{63.26} & \multicolumn{1}{c|}{1.34}  & \multicolumn{1}{c|}{3.83}  & \multicolumn{1}{c|}{2.59}  & \multicolumn{1}{c|}{}                  & \multicolumn{1}{c|}{3.43$\pm$2.73}   & \multicolumn{1}{c|}{2.33$\pm$1.91}  & \multicolumn{1}{c|}{2.88$\pm$2.23}  & \multicolumn{1}{c|}{}                  & \multicolumn{1}{c|}{3.39}  & \multicolumn{1}{c|}{0.87}  & 2.13  \\ \hline
		\end{tabular}
	}
	\label{tab:flexible_results}
\end{table*}

{ \textbf{Cross-Testing Results.}}
Despite our model achieving better performance on multi-modal datasets compared to the baselines, it is crucial to check the generalization of the models by evaluating transfer performance on other multi-modal and single-modal datasets. In this section, we conduct cross-testing experiments that models are trained on Protocol ``seen'' of WMCA, and tested on MmFA, Protocol 1 of OULU~\cite{Boulkenafet2017OULU} and SiW~\cite{Liu2018Learning} datasets, respectively. To demonstrate the contribution of each improvement to the generalization and modal invariance, we report the baseline method ViT, two variants MMA-ViT and MFA-ViT, and intra-testing results on WMCA. The results are listed in Tab.~\ref{tab:ablation_results}.
\begin{table}[]
	\centering
	\caption{The results from the intra-testing (ACER (\%)) and cross-testing (HTER (\%)) when models are trained on WMCA, and tested on WMCA, MmFA, OULU and SiW datasets.
	}
	\resizebox{1.0\linewidth}{!}{
		\begin{tabular}{|c|ccc|ccc|c|c|}
			\hline
			\multirow{2}{*}{Method} & \multicolumn{3}{c|}{\begin{tabular}[c]{@{}c@{}}Tested on\\ WMCA\end{tabular}}           & \multicolumn{3}{c|}{\begin{tabular}[c]{@{}c@{}}Tested on\\ MmFA\end{tabular}}             & \begin{tabular}[c]{@{}c@{}}Tested on\\ OULU\end{tabular} & \begin{tabular}[c]{@{}c@{}}Tested on \\ SiW\end{tabular} \\ \cline{2-9} 
			& \multicolumn{1}{c|}{R}             & \multicolumn{1}{c|}{D}             & I             & \multicolumn{1}{c|}{R}              & \multicolumn{1}{c|}{D}             & I              & R                                                        & R                                                        \\ \hline
			ViT                     & \multicolumn{1}{c|}{7.33}          & \multicolumn{1}{c|}{3.68}          & 4.66          & \multicolumn{1}{c|}{27.56}          & \multicolumn{1}{c|}{12.61}         & 37.40          & 45.99                                                    & 42.16                                                    \\ \hline
			MFA-ViT                 & \multicolumn{1}{c|}{6.06}          & \multicolumn{1}{c|}{4.05}          & 5.24          & \multicolumn{1}{c|}{27.86}          & \multicolumn{1}{c|}{18.47}         & 35.48          & 42.61                                                    & 39.32                                                    \\ \hline
			MMA-ViT                 & \multicolumn{1}{c|}{4.78}          & \multicolumn{1}{c|}{3.13}          & 3.51          & \multicolumn{1}{c|}{27.53}          & \multicolumn{1}{c|}{15.79}         & 34.45          & 34.43                                                    & 31.66                                                    \\ \hline
			FM-ViT                  & \multicolumn{1}{c|}{\textbf{2.87}} & \multicolumn{1}{c|}{\textbf{2.32}} & \textbf{2.13} & \multicolumn{1}{c|}{\textbf{26.71}} & \multicolumn{1}{c|}{\textbf{8.01}} & \textbf{31.23} & \textbf{30.70}                                & \textbf{29.98}                                           \\ \hline
		\end{tabular}
	}
	\label{tab:ablation_results}
\end{table}

We can observe that our FM-ViT achieves the best results in both intra-testing experiments on WMCA dataset and cross-testing experiments on OULU and SiW datasets. For example, on the RGB, Depth and IR modalities of WMCA dataset, the ACER of our FM-ViT are $2.87\%$, $2.32\%$ and $2.13\%$ respectively. On the OULU and SiW dataset, the HTER are $30.70\%$ and $29.98\%$ respectively. By comparing MFA-ViT, MMA-ViT and FM-ViT, we can conclude that the benefit of generalization mainly benefits from MMA module, which is consistent with the fixed modal test scenarios.

\subsection{Ablation Study}
{ \textbf{Effect of the Frameworks.}}
We conduct an in-depth analysis of the framework with a variety of ViT family architectures, \ie, baseline method ViT, ``tiny'' and ``base'' variants FM-ViT(T) and FM-ViT(B), and some variants based on FM-ViT(S) framework, \ie, FM*-ViT(S) means changing the order of MMA and MFA in CMTB, $K=4$ indicates that the transformer encoder contains 4 ``Stages'', in which each ``Stage'' contains 2 STBs to ensure that the total length unchanged, and $L=2$ indicates that one ``Stage'' contains 2 CMTBs. Besides ACER, FLOPs and parameters are also employed for comparisons as shown in Tab.~\ref{tab:ablation_frameworks}. Compared with the baseline method ViT, \textbf{our approach (FM-ViT(S))} only introduces 0.80 (G) FLOPs and 7.43 (M) parameters, but has significant performance improvement in three test modalities. Changing the order of MMA and MFA in each CMTB will result in performance degradation. We analyze that the efficiency of modal fusion will be reduced if it is executed before informative feature mining. Finally, whether adding CMTB by stacking more ``Stage'' (see K=4) and increasing the number of CMTB in each ``Stage'' (see L=2), there is no performance benefit and more FLOPs and parameters are introduced.

Based on the ViT(S), we will further improve the performance by replacing BCE with CMFL~\cite{george2021cross}, which aims to modulate the loss contribution of each channel as a function of the confidence of individual channels. For example, the ACER of RGB, Depth, IR modalities is reduced from $4.32\%$, $2.83\%$, $2.34\%$ to $4.10\%$, $2.80\%$, $2.22\%$, respectively. Similar conclusion on the FM-ViT show that CMFL can be used as an independent transferable loss to replace BCE in multi-branch architecture.
\begin{table}[]
	\centering
	\caption{Ablation study with different architecture on WMCA with Protocol ``seen''. The \textcolor{red}{red} color indicates changes from FM-ViT(S). $K$, $M$ and $L$ denote the number of ``Stage'' in transformer encoder, the number of STB and CMTB in one ``Stage''.}
	\resizebox{1.0\linewidth}{!}{
		\begin{tabular}{|l|c|c|c|c|c|c|}
			\hline
			\multicolumn{1}{|c|}{Model} & K & {[}$M^{1}$ $\sim$ $M^{K}${]}   & L & \begin{tabular}[c]{@{}c@{}}ACER\\ (R/D/I)\end{tabular} & \begin{tabular}[c]{@{}c@{}}FLOPs\\ (G)\end{tabular} & \begin{tabular}[c]{@{}c@{}}Params\\ (M)\end{tabular} \\ \hline
			\textcolor{red}{ResNet50}    & - & -         & - & \textcolor{red}{10.12/6.93/5.85}  & \textcolor{red}{4.10}     & \textcolor{red}{25.00}   \\ \hline
			\textcolor{red}{ViT(S)}    & - & -         & - & 4.32/2.83/2.34   & 3.02       & 15.34                                                \\ \hline
			\textcolor{red}{ViT(S) w/ CMFL}    & - & -         & - & 4.10/2.80/2.22                                         & 3.02  & 15.34 \\ \hline
			FM-ViT(\textcolor{red}{T})     & 3 & [4, 4, 4] & 1 & 6.06/4.05/5.24                        & 0.39                                                & 2.17  \\ \hline
			\textbf{FM-ViT(S)}           & 3 & [2, 2, 4] & 1 & 2.87/2.32/2.13  & 3.85                         & 22.77                           \\ \hline
			\textcolor{red}{FM-ViT w/ CMFL}           & 3 & [2, 2, 4] & 1 & 2.85/2.30/2.06    & 3.85  & 22.77    \\ \hline
			FM-ViT(\textcolor{red}{B})                   & 3 & [4, 4, 4] & 1 & 6.99/7.22/6.87                                         & 6.09                                                & 34.12                                                \\ \hline
			FM\textcolor{red}{*}-ViT(S)   & 3 & [2, 2, 4] & 1 & 3.13/3.51/1.91       & 3.85        & 22.77                 \\ \hline
			FM-ViT(S)                   & \textcolor{red}{4} & [2,2,2,2] & 1 & 3.22/3.38/4.91                                         & 4.13                                                & 25.25                                                \\ \hline
			FM-ViT(S)                   & 3 & [2, 2, 4] &  \textcolor{red}{2} & 6.98/3.24/5.81                                         & 4.69                                                & 30.21                                                \\ \hline
		\end{tabular}
	}
	\label{tab:ablation_frameworks}
\end{table}

\begin{figure}[t]
	\begin{minipage}{0.49\linewidth}
		\centerline{\includegraphics[width=1\textwidth]{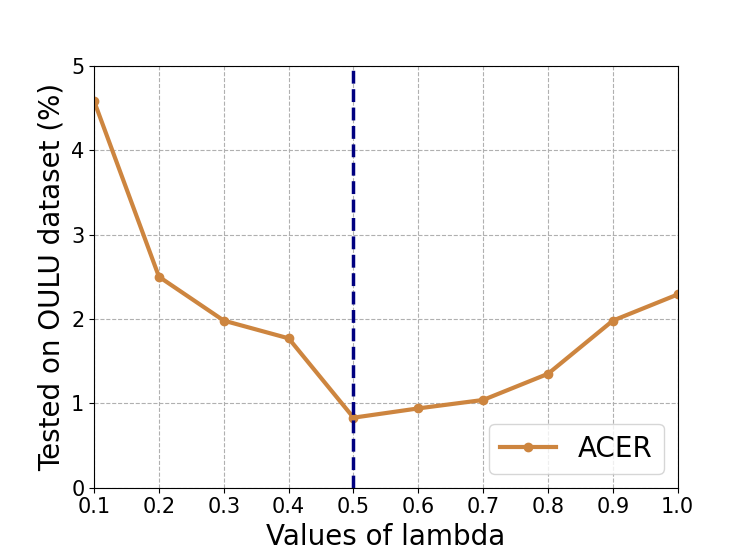}}
		\centerline{(a)}
	\end{minipage}
	\
	\begin{minipage}{0.49\linewidth}
		\centerline{\includegraphics[width=1\textwidth]{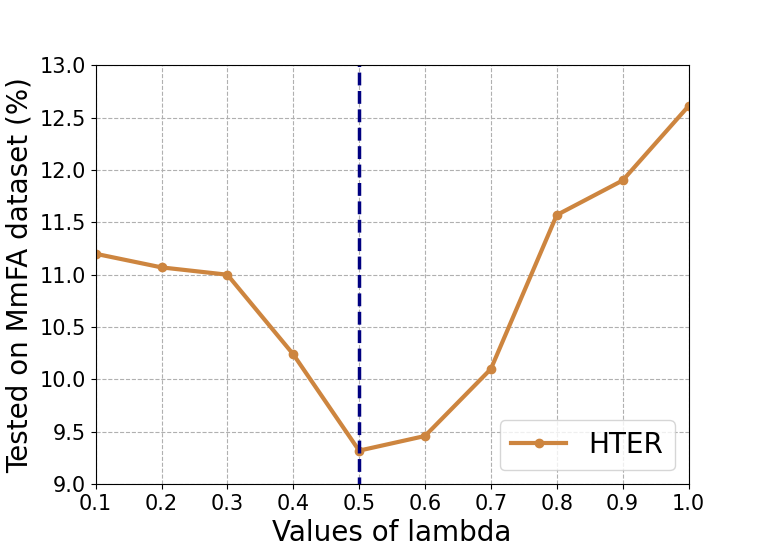}}
		\centerline{(b)}
	\end{minipage}
	\vspace{-0.2cm}
	\caption{(a) Intra-testing experiment. The results of ACER on OULU with different $\lambda$ values. (b) Cross-testing experiment. The results of HTER on MmFA with different $\lambda$ values.}
	\label{fig_parameter}
\end{figure}

{ \textbf{Effect of the $\lambda$ in MMA.}}
The threshold function $\Gamma_{\lambda }(\cdot )$ selects the informative patch tokens in MMA according to the parameter $\lambda$. We conduct experiments on MMA-ViT with both intra-testing and cross-testing experiments to search a best value for $\lambda$. The model is trained with Protocol 1 of OULU dataset for intra-testing and trained with Protocol ``seen'' of WMCA and tested on MmFA dataset for cross-testing. As shown in Fig.~\ref{fig_parameter}, all experiments achieve the best results when setting $\lambda=0.5$, which means we select informative patch tokens by thresholding the attention map to keep 50\% of the mass.

\subsection{Visualization Analysis}
\begin{figure}[t]
	\centering
	\includegraphics[width=1.0\linewidth]{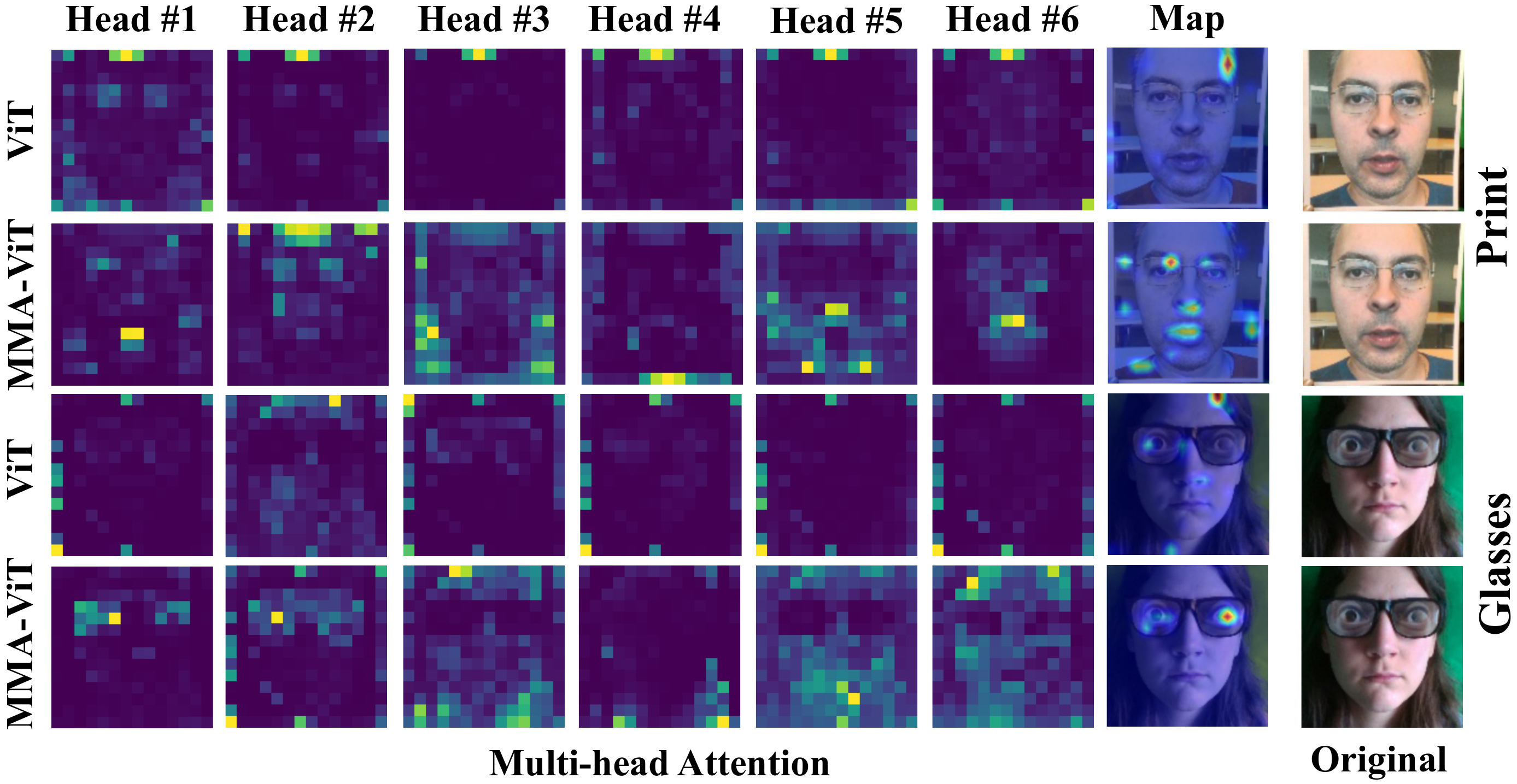}
	\caption{Comparison between multi-head attention visualization on the last layer of vanilla ViT and MMA-ViT. Images are selected from WMCA.}
	\label{fig:MMA-ATTN-MAP}
\end{figure}

\begin{figure}[t]
	\centering
	\includegraphics[width=1.0\linewidth]{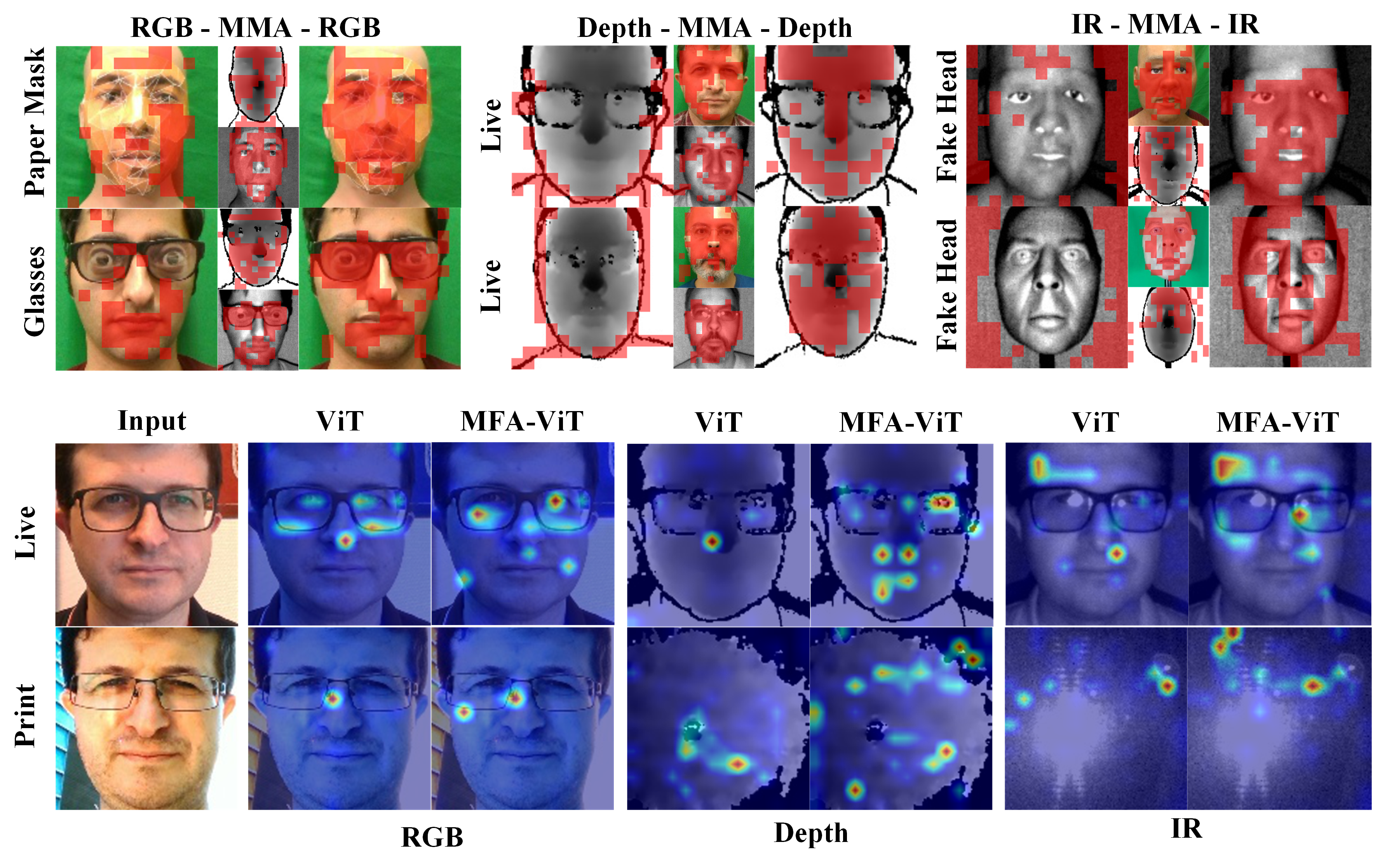}
	\vspace{-0.4cm}
	\caption{(a) Top: the mining process (``before-MMA-after'') of informative patch tokens in each modal sequence which are covered by the red mask and obtained by thresholding the MA maps to keep 50\% ($\lambda=0.5$) of the mass~\cite{caron2021emerging}. (b) Bottom: feature map visualization using Transformer Attribution method~\cite{chefer2020transformerInterpretability}.}
	\label{fig:mma_mfa_vis}
\end{figure}

In Fig.~\ref{fig:mma_mfa_vis} (a), we visualize the patch tokens mined by the MMA in the last ``Stage''. For each modal branch in the MMA-ViT model, we visualize the areas of informative patch tokens before and after a randomly selected head. Such as for RGB input, two attack samples from the Protocol ``seen'' of WMCA test set are shown in Fig.~\ref{fig:mma_mfa_vis} (a). We can see that before MMA, the informative patch tokens are not distributed in the areas with obvious spoofing clues, \ie, the nose position of ``Paper Mask'' and the eye position of ``Glasses''. While, the distribution is adjusted after MMA, which refers to the informative patch tokens in other modal sequences to mine potential patch tokens ignored by RGB sequences. In Fig.~\ref{fig:MMA-ATTN-MAP}, we further visualize the response area of the last layer's six multi-head attention. The example of Print attack indicates that the attention regions of ViT’s heads overlap to a large extent, while our proposed MMA-ViT has four heads (head $\#$1, 2, 5, 6) focus on the mouth, forehead, nose and depth information of the face, and two heads (head $\#$3, 4) attend to the boundary information of the printed paper. A similar phenomenon applies to funny eyes glasses, such as three heads (head $\#$1, 2, 6) fortunately paid attention to the deliberate occlusion.

In Fig.~\ref{fig:mma_mfa_vis} (b), we verify the ability of MFA to absorb multi-modal information. The comparison models are trained on RGB, Depth, and IR modalities with baseline ViT. We randomly select the samples correctly classified by the baseline model for visualization, to ensure that class-dependent relevance maps are generated. We can observe from Fig.~\ref{fig:mma_mfa_vis} (b) that whether it is a live face or an attack sample, the highlight areas (high relevancy scores) output by MFA-ViT is expanded on the baseline method ViT. It means that MFA interacts with other modal sequences with the help of $\texttt{CLS}$ token, and further enriches each patch token with the fused multi-modal information.

\section{Conclusion}
In this work, we present a pure transformer-based framework named FM-ViT to improve the performance of a single-modal FAS system with the help of multi-modal data. FM-ViT introduces CMTB after some specific STBs, which consists of two cascaded attentions named MMA and MFA to guide each branch to learn potential and modality-agnostic liveness features, respectively. Experiments show that our approach FM-ViT only introduces 0.80 (G) FLOPs and 7.43 (M) parameter for the ``small'' variant, but gains significant improvements compared with the baseline method ViT.

%\section{Acknowledgements}
%This work was supported by the Chinese National Natural Science Foundation Projects $\#$61961160704, $\#$61876179, the External cooperation key project of Chinese Academy Sciences $\#$ 173211KYSB20200002, the Key Project of the General Logistics Department Grant No.AWS17J001, Science and Technology Development Fund of Macau (No.~0025/2018/A1, 0010/2019/AFJ, 0025/2019/AKP). \emph{(Ajian~Liu and Zichang~Tan are joint first authors.) (Corresponding author: Jun~Wan and Yanyan~Liang.)}

\ifCLASSOPTIONcaptionsoff
  \newpage
\fi
\bibliographystyle{IEEEtran}
\bibliography{IEEEabrv,reference}

\end{document}